\documentclass{article}

\usepackage{microtype}
\usepackage{graphicx}
\usepackage{subfigure}
\usepackage{booktabs} 
\usepackage[ruled]{algorithm2e}
\usepackage{diagbox}
\usepackage{hyperref}

\usepackage[accepted]{icml2021}
\usepackage{comment}
\usepackage{bm}
\usepackage{amssymb}
\usepackage{amsmath}
\usepackage{color}
\usepackage{multirow}
\usepackage{multicol}
\usepackage{dsfont}

\newcommand{\qileft}{[\kern-0.15em[}
\newcommand{\qiLeft}{\left[\kern-0.4em\left[}
\newcommand{\qiright}{]\kern-0.15em]}
\newcommand{\qiRight}{\right]\kern-0.4em\right]}
\DeclareMathOperator*{\Exp}{\mathbb{E}}

\renewcommand{\a}{{\bm{a}}}

\renewcommand{\c}{{\bm{c}}}

\renewcommand{\r}{{\bm{r}}}
\newcommand{\s}{{\bm{s}}}

\newcommand{\w}{{\bm{w}}}

\newcommand{\bsigma}{{\bm{\sigma}}}
\newcommand{\blambda}{{\bm{\lambda}}}

\newcommand{\btheta}{{\bm{\theta}}}
\newcommand{\bLambda}{{\bm{\Lambda}}}

\newcommand{\bw}{{\bm{W}}}

\newcommand{\bTheta}{{\bm{\Theta}}}

\newcommand{\A}{{\mathcal{A}}}

\newcommand{\D}{\mathcal{D}}

\newcommand{\N}{\mathcal{N}}

\newcommand{\wrt}{{{w.r.t.}}}
\newcommand{\ie}{{\emph{i.e.}}}
\newcommand{\eg}{{\emph{e.g.}}}

\newcommand{\st}{\mbox{s.t.}}
\newcommand{\tabincell}[2]{\begin{tabular}{@{}#1@{}}#2\end{tabular}}

\icmltitlerunning{K-shot NAS: Learnable Weight-Sharing for NAS with K-shot Supernets}

\begin{document}

\twocolumn[
\icmltitle{K-shot NAS: Learnable Weight-Sharing for NAS with K-shot Supernets}



\icmlsetsymbol{equal}{*}

\begin{icmlauthorlist}
\icmlauthor{Xiu Su}{equal,usyd}
\icmlauthor{Shan You}{equal,sensetime,thu}
\icmlauthor{Mingkai Zheng}{sensetime}
\icmlauthor{Fei Wang}{sensetime}
\icmlauthor{Chen Qian}{sensetime}
\icmlauthor{Changshui Zhang}{thu}
\icmlauthor{Chang Xu}{usyd}
\end{icmlauthorlist}

\icmlaffiliation{usyd}{School of Computer Science, Faculty of Engineering, The University of Sydney, Australia}
\icmlaffiliation{sensetime}{SenseTime Research}
\icmlaffiliation{thu}{Department of Automation, Tsinghua University,
Institute for Artificial Intelligence, Tsinghua University (THUAI), 
Beijing National Research Center for Information Science and Technology (BNRist)}

\icmlcorrespondingauthor{Shan You}{youshan@sensetime.com}

\icmlkeywords{Machine Learning, ICML}

\vskip 0.3in
]



\printAffiliationsAndNotice{\icmlEqualContribution} 

\begin{abstract}
In one-shot weight sharing for NAS, the weights of each operation (at each layer) are supposed to be identical for all architectures (paths) in the supernet. However, this rules out the possibility of adjusting operation weights to cater for different paths, which limits the reliability of the evaluation results. In this paper, instead of counting on a single supernet, we introduce $K$-shot supernets and take their weights for each operation as a dictionary. The operation weight for each path is represented as a convex combination of items in a dictionary with a simplex code. 
This enables a matrix approximation of the stand-alone weight matrix with a higher rank ($K>1$). A \textit{simplex-net} is introduced to produce architecture-customized code for each path. As a result, all paths can adaptively learn how to share weights in the $K$-shot supernets and acquire corresponding weights for better evaluation.
$K$-shot supernets and simplex-net can be iteratively trained, and we further extend the search to the channel dimension. Extensive experiments on benchmark datasets validate that K-shot NAS significantly improves the evaluation accuracy of paths and thus brings in impressive performance improvements.


\end{abstract}

\begin{figure*}[t]
    \centering
    \includegraphics[width=0.95\linewidth]{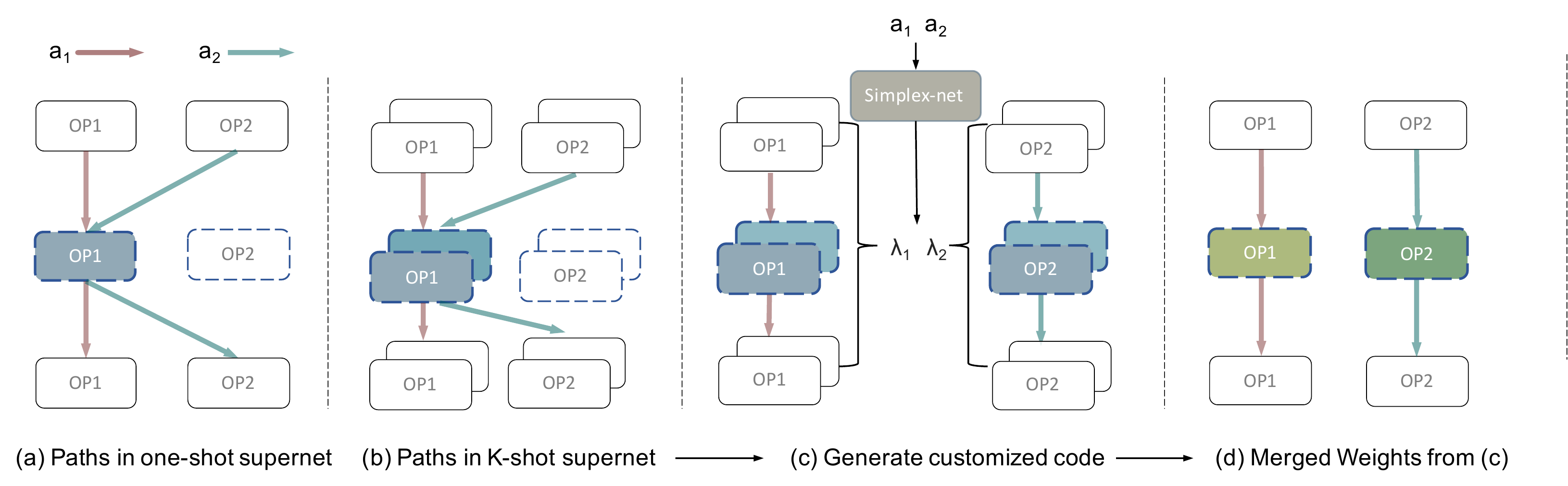}
    \caption{K-shot NAS directly learns the customized code (coefficient vector) $\blambda$ for different paths (subnets or architectures) $\a$.  Therefore, even with the same operation, \eg~OP1 in the second layer, the weights of each path is encouraged to be different from each other as in (d) and be a more accurate approximation for the stand-alone weights trained from scratch. However, as in (a), paths in one-shot supernet fully share a same set of weights even when architectures are different ($\a_1$ vs $\a_2$).}
    \label{motivation}
\end{figure*}
\section{Introduction}
Deep neural models have thrived in various walks of life with artificial intelligence and edge computing. Before deploying a model for a task of interest, practitioners need to consider the hardware budgets and specify a decent architecture or structure for the model. In this way, much effort have devoted to designing sophisticated structures, such as ResNet \cite{he2016deep}, SENet \cite{hu2018squeeze} and MobileNet \cite{mobilenetv2}, and further boosting the performance within the hardware budgets, \eg, knowledge distillation \cite{hinton2015distilling,you2017learning,du2020agree} and quantization \cite{gholami2021survey}. Recently, neural architecture search (NAS) \cite{mnasnet,fbnetv2,guo2020single} merges both sides and aims to directly generate architectures with a promising performance by liberating manual labor and surmounting the cognitive bottleneck of humans.



Pioneer works usually design architectures via a decision-making process with reinforcement learning \cite{zoph2016neural} or pure evolutionary algorithms \cite{real2019regularized}. However, they often involve unbearable training costs and resource consumption. For the sake of searching efficiency, current methods mainly leverage a one-shot supernet with all architectures (paths) that can be derived and share their weights mutually. Gradient-based methods introduce architecture parameters to deferentially
optimize the weights and architecture parameters, such as DARTS \cite{liu2018darts} and its variants \cite{xu2019pc,huang2020explicitly,yang2020ista,yang2021towards}. But they may suffer the significant memory cost and the detrimental gap when deriving the final architecture. Sample-based methods \cite{you2020greedynas} thus propose to decouple the supernet weights and architecture parameters. They train the supernet by sampling paths from it \cite{guo2020single} and then take it as an evaluator for ranking different paths, with the more stable and promising performance achieved.



In one-shot NAS methods, the assumption of weight sharing plays an important role. Specifically, it postulates that the weights of an operation at a layer are identical for all the paths containing this operation. Thus, the training efficiency of the supernet can be largely ensured, as only one copy of the operation weight has to be maintained for the massive number of paths. However, this also rules out the possibility of adjusting the operation weight to cater for different paths. As a result, any two different paths will always be compulsively required to inherit the weights of the same operations. Nevertheless, they may be far from the \textit{stand-alone} weights obtained by training each path from scratch. With such a harsh and limited space to adapt to the optimal operation weight, the resulting evaluation performance of paths could be unreliable and rarely reflects the Oracle ranking of different paths for supernet.

In this paper, we introduce a K-shot NAS framework to unleash the potential of sampled subnetworks (subnets). Instead of counting on a single operation weight of supernet, we maintain a dictionary of weights for an operation based on a group of $K$ supernets (see Figure \ref{motivation}). All paths containing a certain operation are supposed to share the dictionary mutually, but we allow them to have their own customized manner to exploit it. Particularly, the operation weight for each path is represented as a convex combination of items in the dictionary with a code on the simplex. We encode the architecture in a code generator \textit{simplex-net} to generate these codes customized for different paths. Then we can interpret the weight sharing as a matrix approximation problem, where the one-shot paradigm merely leverages a rank-1 matrix to approximate the stand-alone weight matrix for all paths while our $K$-shot case encourages the approximation through a matrix of higher rank.  We further consider the joint searching with channel dimension (number of filters for operations). A channel branch can be encoded into the simplex-net to explore the context information or the difference among layers. We propose a non-parametric regularization to improve the discrimination between simplex codes of different widths. Extensive experiments on the benchmark ImageNet \cite{russakovsky2015imagenet} and NAS-Bench-201 \cite{nasbench201} datasets validate the effectiveness of our $K$-shot NAS and advantages over other one-shot NAS methods. We can achieve a 77.9\% Top-1 accuracy with the MobileNetV2 search space with only 412M FLOPs.

\section{Rethinking Weight-Sharing with Supernets}
One-shot NAS methods mainly dedicate to optimize a supernet $\mathcal{N}$ which contains all the architectures from a given search space $\mathcal{A}$. 
The weight $\btheta$ of the supernet $\mathcal{N}$ is shared by all its sub-networks (a.k.a architectures or paths), \ie, $\a \in \mathcal{A}$.
With the trained supernet, the optimal architectures can be identified through searching and evaluating them. 
Following \cite{guo2020single,you2020greedynas}, a general one-shot NAS can be formulated as a two-stage optimization problem, \ie, weights optimization for supernets and architecture search, 
\begin{align}\label{eq1}
\bm{a}^* &= \mathop{\arg\max}_{\a \in \A}~\mathop{\mbox{ACC}_{val}}(\a, \btheta^*_{\A}(\a))
\end{align}
\begin{align}\label{eq1}
\st~\btheta^*_{\A} &= \mathop{\arg\min}_{\btheta}~ \mathcal{L}_{train}(\A,\btheta),
\end{align}
where $\mbox{ACC}_{val}$ denotes the accuracy for each architecture $\a$ on the validation dataset $\D_{val}$, and $\mathcal{L}_{train}(\A,\btheta)=\Exp_{\bm{a} \in \mathcal{A}}\qiLeft \mathcal{L}_{train}(\a,\btheta) \qiRight$ denotes the training loss for the supernet on the training dataset $\D_{train}$ by randomly sampling a path and optimizing its weights accordingly. 
Since it is computationally prohibited to traverse all architectures in a large search space, \eg, $13^{21}$ possible networks in a MobileNetV2 search space, the optimal architecture search in Eq.\eqref{eq1} usually turns to efficient searching algorithms, such as evolutionary search \cite{deb2002fast,guo2020single,you2020greedynas}.

\subsection{One-shot Weight Sharing}
One-shot supernet adopts a fixed weight-sharing paradigm. That is, the weights of an operation at a layer are supposed to be identical (fully-shared) for all sampled architectures that contain this operation. Consider a supernet $\N$ with $L$ layers and $O$ operations for each layer, and then the overall search space size is $O^{L}$. Formally, in a one-shot weight sharing strategy, for a certain operation $o$ at a layer, then it will be involved in $N=|\A|/O = O^{L-1}$ possible paths, which we simply denote as $\{\a_1,...,\a_N\}$. Besides, if we train all these $N$ paths from scratch, we can have their Oracle or \textit{stand-alone} weights on the operation $o$, \ie, $\bw = [\w_1,...,\w_N]\in\mathbb{R}^{d\times N}$.   As for the one-shot paradigm, all weights for different paths will amount to the same weight $\btheta_o$ of the supernet, and the stand-alone matrix $\bw$ is assumed to be approximated as  
\begin{equation}\label{sharing:oneshot}
\mbox{one-shot:} \quad \bw = [\w_1,...,\w_N] \approx \bm{1}^T \otimes \btheta_o,
\end{equation}
where $\btheta_o$ represents the weight of operation $o$ in the supernet, and $\otimes$ is the Kronecker product between two vectors with the vector $\bm{1}\in\mathbb{R}^{N}$ being all ones. The one-shot weight sharing can largely benefit the training efficiency of the supernet, as only one copy of the operation weight has to be maintained for the massive number of architectures. However, the one-shot weight sharing in Eq.\eqref{sharing:oneshot} rules out the possibility of adjusting the operation weight to cater to different paths. For any two paths $\a_i,\a_j$ that involve the same operation $o$ from the supernet, they will always share the same weight of $o$. With such a limited room to match a satisfactory operation weight, the resulting evaluation performance of the paths could be unconvincing and hardly reflects the potential of different paths. 
\subsection{K-shot Weight Sharing}

Instead of relying on the single operation weight $\btheta_o$ from the supernet, we tend to maintain a dictionary of weights for an operation. Suppose there are $K$ supernets $\{\btheta_1, \cdots, \btheta_k\}$. In terms of the operation $o$, we can set up a dictionary of weights $\bTheta=\{\btheta^1_o, \cdots, \btheta^K_o\}$ based on these $K$ supernets. For simplicity, we will omit the subscript $o$ in the sequel. The dictionary $\bTheta$ can be shared by different paths that contain the operation $o$, but we allow them to have their own way to exploit the dictionary of operation weights. In particular, for the $i$-th path, we calculate the operation weight through a convex combination of items in the dictionary,
\begin{equation} \label{approxmation_weights}
 \w_i\approx \bTheta\blambda_i= \sum_{k=1}^K\lambda_{i,k}\btheta_k,   
\end{equation}
where $\blambda_i$ stands for the combination code of the $i$-th path and lies on the simplex $\Delta^{K-1}$, 
\begin{equation*}\label{sharing:simplex}
\Delta^{K-1}=\{\blambda|\sum_{k=1}^K\lambda_k=1, \lambda_k\geq0,~\forall~k\in \{1, \cdots, K\}\}.
\end{equation*}
By simultaneously considering all $N$ paths that involve the operation $o$, we can approximate their operation weights as
\begin{equation}\label{sharing:kshot}
 	\mbox{$K$-shot:} \quad \bw = [\w_1,...,\w_N] \approx \bTheta \bLambda,
\end{equation}
where $\bTheta$\footnote{Since $\bTheta$ indicates the weights of any operation from $K$ supernets. Without loss of generality, we also use $\bTheta$ to indicate the weights of $K$ supernets $\{\btheta_1, \cdots, \btheta_k\}$ accordingly.} serve as a weight basis for all paths, and $\bLambda$ indicates the customized code for paths $\{\a_1,...,\a_N\}$. In this way, all architectures can inherit the weights from the same set of $K$ supernets simultaneously, yet with their associated code $\blambda$. We thus name this paradigm \textbf{\textit{$K$-shot weight-sharing}}.
\begin{figure}[t]
\label{fig:framework}
	\centering
	\includegraphics[width=0.95\linewidth]{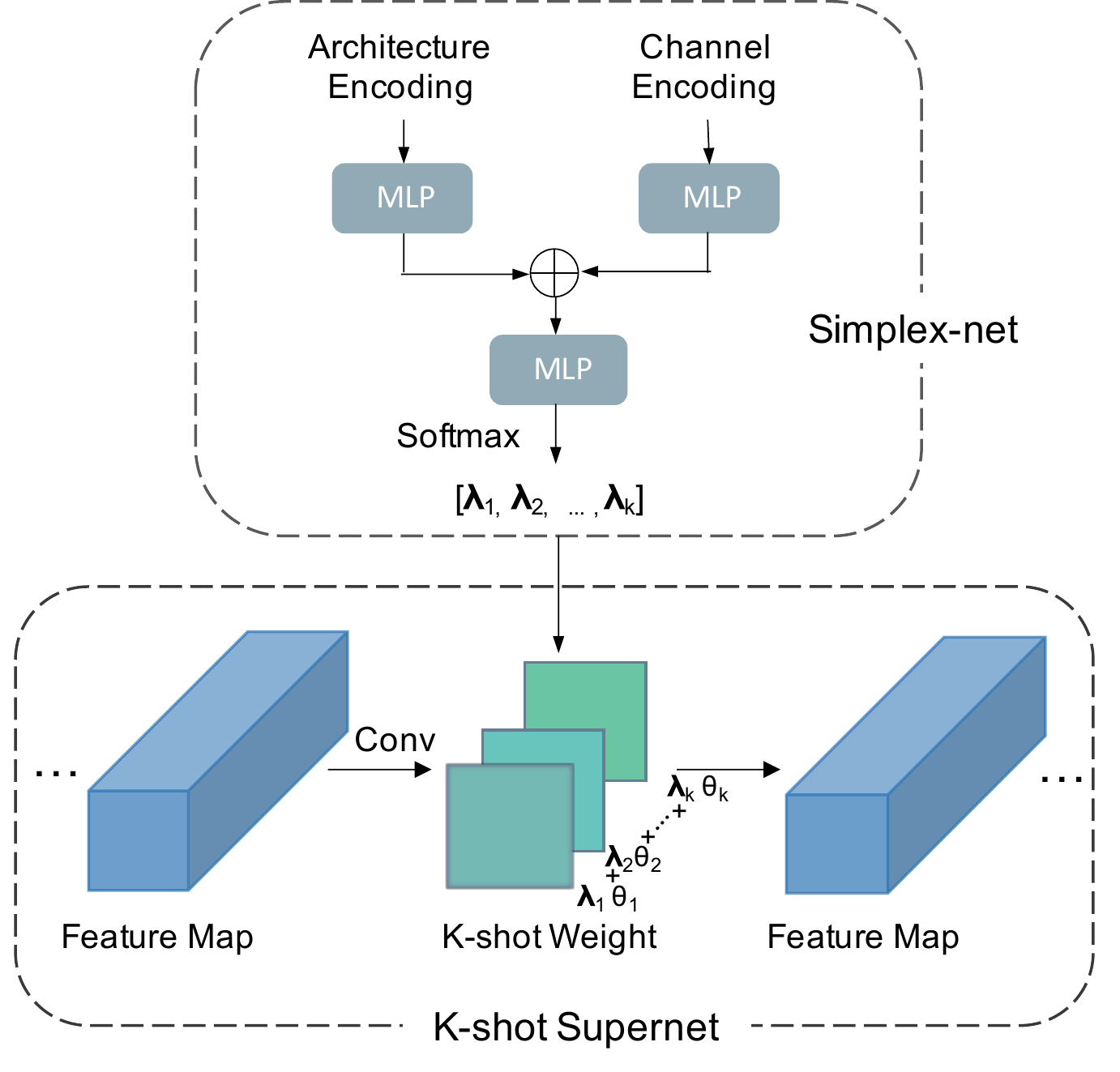}
	\caption{The framework of simlex-net and $K$-shot Supernet. With simplex-net, each architecture learns the customized code for constructing new supernets with $K$-shot weight, thus bridging the gap between supernets and stand-alone training. }
 	\vspace{-1.5em}
\end{figure}

Actually, the weight sharing paradigm can be interpreted as a matrix approximation problem \cite{tropp2017practical}. As for one-shot case in Eq.\eqref{sharing:oneshot}, it leverages a rank-1 matrix $\bm{1}^T \otimes \w$ to approximate the optimal stand-alone weight matrix $\bw$ for various paths. In contrast, we encourage an approximation of $\bw$ through a bit higher rank matrix. Since $d\ll N$ in the size of $\bw$, we can leverage a rank $K~(1\leq K\leq d)$ to fulfill the approximation, and it can be written as the multiplication of two small matrices, $\bTheta\in\mathbb{R}^{d\times K}$ and $ \bLambda\in\mathbb{R}^{K\times N}$ in Eq.\eqref{sharing:kshot}. By dint of the low-rank approximation \cite{de2008tensor,kishore2017literature,clarkson2017low}, the gap between the ideal weight matrix and its approximation can be reduced compared to the one-shot paradigm. The classical rank-1 assumption is thus taken as a special case of ours. Note that the approximation in Eq.\eqref{sharing:kshot}  depends on the optimality of the combination code $\bLambda$. If all combination code $\blambda$ are nearly the same, then $K$-shot supernets will degrade to a one-shot case with $\w = \bTheta\blambda$. We thus hope the combination code of paths can be different and customized to reflect the characteristics of their architectures.

\newcommand{\norm}[1]{\left\lVert#1\right\rVert}

\section{NAS with K-shot Supernets} 

As illustrated before, with $K$-shot supernets, weights of paths can have a better approximation to the ideal stand-alone weight matrix than one-shot paradigm, thus promoting the evaluation of paths to reflect their actual potential. Formally, by examining Eq.\eqref{approxmation_weights}, each code $\blambda_i$ can be seen as a point on the simplex to weigh the $K$ supernets.  The weight of an operation shall not be only determined by itself (chosen or not). Instead, the architecture $\a$ of the subnets as a kind of context information should influence the optimal weight of an individual operation. We thus re-parameterize $\blambda_i$ by architecture $\a_i$ explicitly via a \textit{simplex-net} $\pi$ on simplex $\Delta^{K-1}$, \ie,
\begin{equation}\label{simplex-net}
\pi: \A\rightarrow \Delta^{K-1}, \quad \a_i\mapsto \pi(\a_i;\bsigma) = \blambda_i,	
\end{equation}
where $\bsigma$ is the parameters of the simplex-net, as shown in Figure \ref{fig:framework}, the simplex-net can be modeled with a simple multi-layer perception (MLP) that takes the one-hot architecture encoding vector $\a$ as the input to generate its customized code $\blambda$.  The objective function of $K$-shot supernets can now be formulated as follows:
\begin{equation}\label{eqx}
\begin{aligned}
&\bm{a}^* = \mathop{\arg\max}_{\a \in \A}~\mathop{\mbox{ACC}_{val}}(\a, \widetilde{\btheta}(\bTheta^*_{\A}, \bsigma^*_{\A}, \a)), \\ 
\st~&\bTheta^*_{\A}, \bsigma^*_{\A} = \mathop{\arg\min}_{\bTheta,\bsigma}~ \mathcal{L}_{train}(\A,\widetilde{\btheta}(\bTheta,\bsigma;\A)),\\
&\widetilde{\btheta}(\bTheta,\bsigma;\A)  = \bTheta\cdot\blambda_\A =\bTheta\cdot \pi(\A;\bsigma),
\end{aligned}
\end{equation}
where $\widetilde{\btheta}=\sum_{k=1}^K\lambda_{k}\btheta_k$ indicates the merged new weights from those of $K$ supernets $\bTheta$ and customized code $\blambda$. $\bTheta^*_{\A}, \bsigma^*_{\A}$ is the optimal weights for the supernet and simplex-net after training. As a result, we can learn how all architectures share their weights with $K$-shot supernets by the learned customized code adaptively. In this way, the approximation gap can be reduced compared to the one-shot paradigm, and the evaluation ability is expected to be boosted.



\textbf{Iterative training of supernets and simplex-net.} Commonly, the supernet is trained by sampling a path and a mini-batch $b$ of images per step and optimizing the corresponding weights. However, with the introduced simplex-net, this end-to-end training might bring in difficulty for the simplex-net since the valid batch size for simplex-net amounts to 1 (one path per step).  
In this way, we propose to use a larger path size for the simplex-net but optimize in an iterative manner. In detail, we choose to optimize the simplex-net with the $K$-shot supernets fixed. Then we randomly sample $m$ paths and optimize the simplex-net accordingly. Besides, to ensure the memory cost, we benchmark the same batch size for the images and divide the image batch into $m$ groups, with $b/m$ images for each path. Then the optimization of simplex-net in Eq.\eqref{eqx} for some iteration $\tau$ evolves into 
\begin{equation}\label{iterative_supners}
\begin{aligned}
&\bsigma^{(\tau+1)}_{\A} = \mathop{\arg\min}_{\bsigma}~ \mathcal{L}_{train}(\A,\widetilde{\btheta}(\bTheta^{(\tau)},\bsigma;\A)),\\
&\widetilde{\btheta}(\bTheta^{(\tau)},\bsigma;\A)   =\bTheta^{(\tau)}\cdot \pi(\A;\bsigma) 
\end{aligned}
\end{equation}
As for the training of supernets, we just optimize the weights of supernet with the image mini-batch by fixing the simplex-net. 
The optimization process is thus: 
\begin{equation}\label{iterative_supners}
\begin{aligned}
&\bTheta^{(\tau+1)}_{\A}= \mathop{\arg\min}_{\bTheta}~ \mathcal{L}_{train}(\A,\widetilde{\btheta}(\bTheta,\bsigma^{(\tau)};\A)),\\
&\widetilde{\btheta}(\bTheta,\bsigma^{(\tau)};\A)   =\bTheta\cdot \pi(\A;\bsigma^{(\tau)}) 
\end{aligned}
\end{equation}

For simplicity, we use the iterative training method by default till we converge to the optimal solution $\bTheta^*$ and $\bsigma^*$. And the searching with $K$-shot supernets is formulated as Eq.\eqref{k_search}.
\begin{equation}\label{k_search}
\begin{aligned}
&\bm{a}^* = \mathop{\arg\max}_{\a \in \A}~\mathop{\mbox{Acc}_{val}}(\a,\widetilde{\btheta}^*), \\ 
~\st~ &\widetilde{\btheta}^*(\bTheta^*,\bsigma^*;\a) = \bTheta^*\cdot\blambda_a= \bTheta^*\cdot\pi(\a;\bsigma^*); \\
\end{aligned}
\end{equation}

\begin{table}[t]
	\centering
	\caption{The training and search cost (GPU hours) of 1 epoch with different $K$ on ImageNet dataset. The $4$-th row reports the GPU usage with the batch size 128 for each GPU.}
	\label{time_cost}
	\small
	\begin{tabular}{c|c|c|c|c|c}
	\hline
	Stage & $K$ = 1 & $K$ = 2 & $K$ = 4 & $K$ = 8 & $K$ = 12 \\ \hline
	Training & 2.27 & 2.28 & 2.32  & 2.37 & 2.43 \\ 
	Search & 0.233 & 0.234 & 0.237 & 0.242 & 0.246 \\ 
	GPU usage & 17.5G & 17.7G & 17.9G & 18.4G & 18.9G \\ \hline
	\end{tabular}	
\end{table}

\textbf{Analysis on Training Cost.}  
Although we introduced a simplex-net to generate the customized code, it only involves several fully connected layers; the additional computational cost is negligible.  Besides, our $K$-shot supernets have $(K - 1)$ times more parameters than the ordinary one-shot weight sharing strategy. Intuitively, we might think the $K$-shot will increase the GPU memory usage and training time a lot, but with Eq.\eqref{sharing:kshot}, the $K$-shot weight is merged before the model forwarding, which means there is no additional feature map being generated during the training process. As shown in Table \ref{time_cost}, from the perspective of GPU memory usage and training speed, the cost of $K$-shot supernets is almost the same as the one-shot method ($K=1$). However, introducing a large number of parameters will make the model very hard to optimize. We may need more training epochs to ensure the model has converged. Please see the experimental ablation studies in Section \ref{exp:ablation_studies}.

\textbf{Search the optimal path with an Evolutionary algorithm.} Since the search space is enormous (\eg, $13^{21}$ for searching operations in MobileNetV2 search space), to boost the search efficiency, we leverage the multi-objective NSGA-II \cite{deb2002fast} algorithm for evolutionary search, which is easy to integrate hard FLOPs constraint. We set the population size as 50 and the number of generations as 20. We randomly select a group of architectures within the target FLOPs as the first generation. Then in each iteration, the top 20 architectures with the highest accuracy are selected as parents to generate new architectures via mutation and crossover. After the search, we only retrain the architecture with the highest accuracy from scratch and report its performance.  More details about evolutionary search are illustrated in supplementary materials.

\begin{figure}[t]
    \label{fig:channel}
    \centering
    \includegraphics[width=\linewidth]{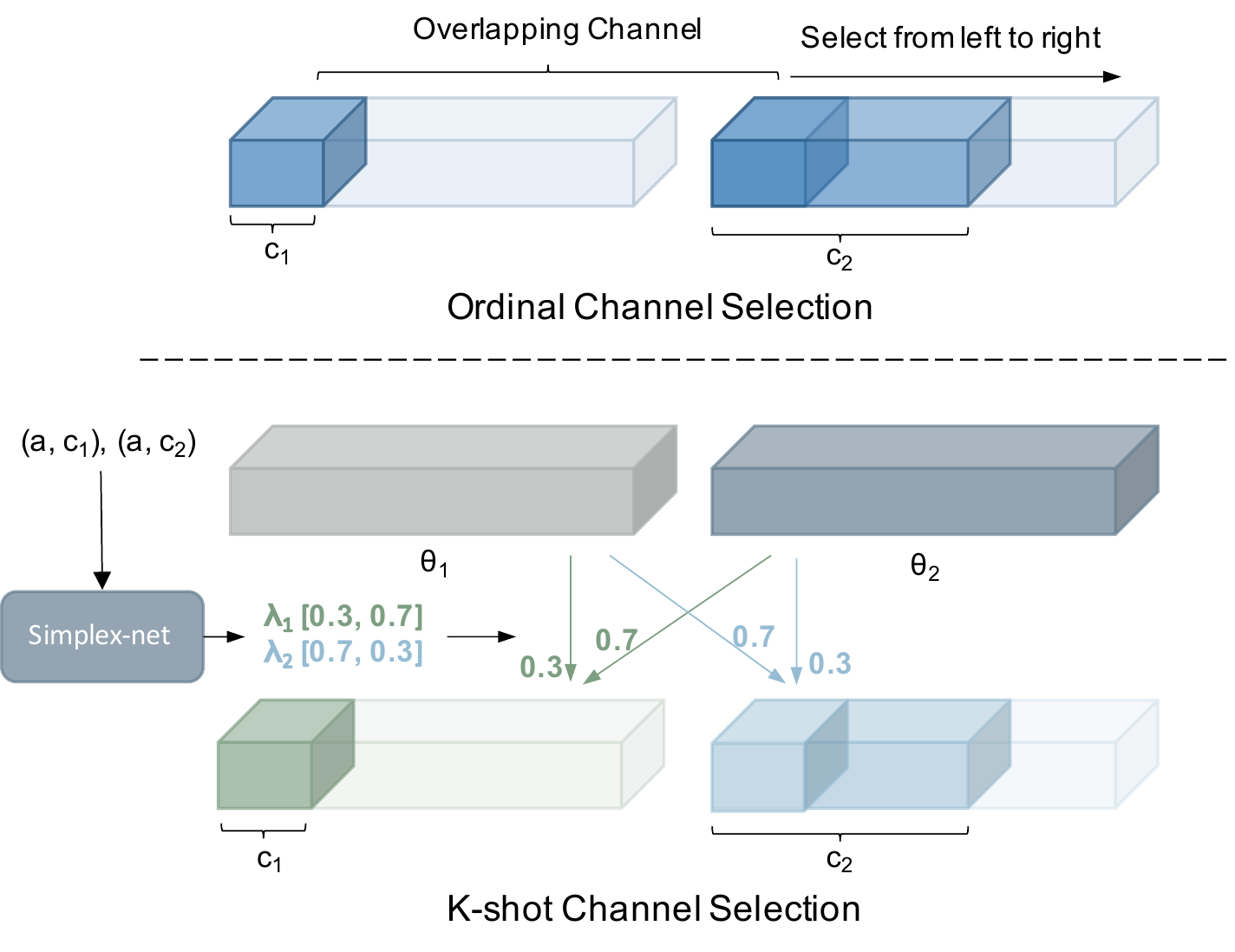}
    \caption{Examples of weight sharing patterns for width. For ordinal weight sharing pattern, leftmost $c$ channels are assigned for the subnets of width $c$. For the width in $K$-shot pattern, each width can customize code for constructing its suitable weight. Therefore, each channel amounts to different weights in $K$-shot pattern.}
    \vspace*{-15pt}
\end{figure}

\section{Joint Searching with Channel Dimension} \label{search_channel}
In previous sections, we have leveraged the $K$-shot supernets and a simplex-net to enable different architectures to have adaptively different weights for some specific operation. Now, we further extend our method to jointly searching with channels for a more fine-grained architecture with decent performance. For clarity, we use $\a$ to denote the architecture (operation) and $\c$ to denote the width structure over all layers, and ($\a$, $\c$) refers to a subnet with architecture $\a$ with width $\c$.

Basically, if we choose different channel width under the same operations, current methods \cite{autoslim,tas,locally,bcnet} mainly follow the ordinary weight sharing pattern, which only takes channels from one side (\eg, left). Specifically, to allocate $c$ channels at a certain layer, an ordinary weight sharing pattern assigns the leftmost $c$ channels as Figure \ref{fig:channel}. Such a hard assignment imposes a strong constraint for different channels; that is, though we choose different channel width, the overlapping channels will always share the same set of weights. We expect to decouple the shared weights when we select different channel width by following our proposed idea. To achieve this, we add an additional ``channel branch" in simplex-net, which takes a channel encoding vector and merged with the ``architecture branch" to generate the final customized code (see Figure \ref{fig:framework}). Now, we can include the ``channel branch" to our objective function:
\begin{equation}\label{train_channels}
\begin{aligned}
&\bTheta^*,\bsigma^* = \mathop{\arg\min}_{\bTheta,\bsigma}~ \Exp_{(\bm{a},\c) \in (\mathcal{A},\mathcal{C})} \qiLeft \mathcal{L}_{train}(\bm{a},\bm{c},\widetilde{\btheta}) \qiRight \\
&~\st~ \widetilde{\btheta}(\bTheta,\bsigma;\a,\c) = \bTheta\cdot\blambda_{\a,\c}= \bTheta\cdot \pi(\a,\c;\bsigma) \\
\end{aligned}
\end{equation}
With the channel search included in our objective, we can rewrite the Eq.\eqref{k_search} as:
\begin{equation}\label{search_channels}
\begin{aligned}
&\bm{a}^*,\bm{c}^* = \mathop{\arg\max}_{(\bm{a},\c) \in (\mathcal{A},\mathcal{C})}~\mathop{\mbox{Acc}_{val}}(\a,\c,\widetilde{\btheta}^*), \\ 
~\st~ &\widetilde{\btheta}^*(\bTheta^*,\bsigma^*;\a,\c) = \bTheta^*\cdot\blambda_{\a,\c}= \bTheta^*\cdot\pi(\a,\c;\bsigma^*) \\
\end{aligned}
\end{equation}

\begin{algorithm}[tbp]
    \caption{Training and search with $K$-shot supernets}
    \label{supp_algorithm2}
    \SetAlgoLined
    \SetKwInOut{Input}{Input}
    \KwIn{$K$-shot supernets with weights $\bTheta$, Simplex-net $\pi$ with weights $\bsigma$, maximum training epochs $T$, warmup epochs $T_w$.}
    Init $\tau$ = 0, $\blambda=[\frac{1}{K},...,\frac{1}{K]}$; \\
    \While{$\tau$ $\leq T$}{
        randomly sample an subnets with architecture $\a$; \\
        \uIf{ $\tau\leq T_w$}{
            train $K$-shot supernets with weights of $\bTheta\blambda$ ;}
        \Else{generate customized code $\blambda=\pi(\a;\sigma)$; \\
       calculate regularization term $r_c$ with Eq.\eqref{non_parametric_term};\\
            Iteratively train $k$-shot supernets and simplex-net $\pi$ with Eq.\eqref{iterative_supners} and Eq.\eqref{train_channels};
        }
    }
    \KwOut{The architecture $\a^*$ searched with Eq.\eqref{search_channels}.}
\end{algorithm}

\textbf{Similarity of architectures with different channel width.}
Despite the simplex-net can directly learn the customized code for different channel encoding vectors, we can still add a regularization term to further help it to distinguish different channel width. 


Suppose we have two set of channels $\c_1$ and $\c_2$ (under the same architecture), the corresponding supernets can be expressed by $\bTheta\blambda_1$ and $\bTheta\blambda_2$ (Eq.\eqref{train_channels} and Eq.\eqref{search_channels}), then we can define the similarity of the two supernets by:
\begin{equation}\label{eqxxxxx}
\begin{aligned}
S((\a, \c_1), (\a, \c_2)) = \blambda_1^T \bTheta^T \bTheta \blambda_2 
= \langle \bTheta^T \bTheta, \blambda_1^T \blambda_2 \rangle
\end{aligned}
\end{equation}
with Cauchy–Schwarz inequality \cite{inequality}:
\begin{equation}\label{eqcz}
\begin{aligned}
S((\a, \c_1), (\a, \c_2)) \leq \norm{\bTheta^T \bTheta}\norm{\blambda_1^T \blambda_2}
\end{aligned}
\end{equation}
therefore, the supremacy for the similarity $S$ lies in the weights from supernets and the inner product of $\blambda_1$ and $\blambda_2$. Since we iteratively update supernets and simplex-net, the first term of  $\norm{\bTheta^T \bTheta}$ are fixed during updating simplex-net. As a result, only $\blambda$ affects their inner product.

\begin{table*}[t]
	\caption{Comparison of searched architectures \wrt~different state-of-the-art NAS methods. Search number means the number of evaluated architectures during searching. $\ddagger$: TPU, $\star$: joint search of operations and channel width.} 
	\label{tbl:sota}
	\centering
	\small
	\setlength{\tabcolsep}{1.5mm}{
	\begin{tabular}{rcccccccc} 
	    \hline
		Methods& \tabincell{c}{Top-1\\ (\%)}&\tabincell{c}{Top-5\\ (\%)} & \tabincell{c}{FLOPs\\ (M)} &\tabincell{c}{Params \\ (M)}&\tabincell{c}{Memory cost}&\tabincell{c}{training cost \\ (GPU days)}&\tabincell{c}{search\\ number}&\tabincell{c}{search cost\\ (GPU days)} \\ \hline
		ShuffleNetV2-1.0 \cite{shufflenetv2} &  69.4& - & 146 & - & - & - & - & - \\
		MbnV2-0.5 \cite{mobilenetv2} &  63.3& - & 150 & 1.3 & - & - & - & - \\
		FBNetV2-F3* \cite{fbnetv2} & 73.2 & - & 126 & - & single path&12& - & - \\
		$K$-shot-NAS-D&\textbf{73.4} &90.7&133 &2.6&single path&12& 1000&$<1$\\
		$K$-shot-NAS-D$^\star$&\textbf{73.7} &90.9&145 &2.8 &single path&12& 1000&$<1$\\ \hline 
		SCARLET-C \cite{chu2019scarletnas}&  75.6&92.6&280 & 6.0 &single path&10& 8400 &12\\ 
		MbnV2-1.0 \cite{mobilenetv2} &  72.0& 91.0& 300 &3.4 & - & - & - & - \\
		MnasNet-A1 \cite{mnasnet}& 75.2 & 92.5& 312 &3.9&single path + RL & $288^\ddagger$& 8000 &-\\
		$K$-shot-NAS-C &\textbf{76.3}&92.6&281& 4.4&single path&12& 1000&$<1$\\
		$K$-shot-NAS-C$^\star$ &\textbf{76.5}&92.7&286& 4.7&single path&12& 1000&$<1$\\ \hline	
		Proxyless-R \cite{proxylessnas}&74.6&92.2&320&4.0&two paths&$15$& 1000 &-\\
		AngleNet \cite{hu2020angle} & 74.2 & - & 325 & - & - & 10 & - & - \\
		SPOS \cite{guo2020single}& 76.2&-&328& - &single path&12&1000&$<1$\\
		MnasNet-A2 \cite{mnasnet}& 75.6 & 92.7 & 340 & 4.8&  single path + RL & $288^\ddagger$ & 8000 & -  \\
		ST-NAS-A \cite{powering} &  76.4 & 93.1 & 326 & 5.2 & single path & - & 990 & -  \\
		SCARLET-B \cite{chu2019scarletnas}&  76.3&93.0&329& 6.5 &single path&10&8400&12\\ 
		GreedyNAS-B \cite{you2020greedynas} & 76.8 & 93.0 & 324 & 5.2 & single path & 7 &1000& $<1$ \\
		MCT-NAS-B \cite{prioritized}& 76.9 &93.4&327 &6.3 &single path&12& 100 &$<1$\\
		FairNAS-C \cite{fairnas}&  76.7&93.3&325& 5.6 &single path&-&-\\ 
		BetaNet-A \cite{fang2019betanas}& 75.9 & 92.8 & 333 & 4.1 &single path& 7 & - & - \\
		$K$-shot-NAS-B&\textbf{77.2} &93.3&332 & 6.2&single path&12& 1000&$<1$\\
		$K$-shot-NAS-B$^\star$&\textbf{77.4} &93.5&343 &6.4 &single path&12& 1000&$<1$\\ \hline  
		AngleNet \cite{hu2020angle} & 76.1 & - & 470 & - & - & 10 & - & - \\
		MnasNet-A3 \cite{mnasnet}& 76.7 & 93.3 & 403 & 5.2& single path + RL & $288^\ddagger$& 8000 &-\\
		EfficientNet-B0 \cite{tan2019efficientnet} & 76.3 & 93.2& 390 & 5.3 &single path& - & - & - \\ 
		$K$-shot-NAS-A&\textbf{77.6} &93.6&422 & 6.5 &single path&12& $1000$&$<1$\\
		$K$-shot-NAS-A$^\star$&\textbf{77.9} &93.8&412 &7.8 &single path&12& $1000$&$<1$\\ \hline
	\end{tabular}}
	\label{nas_experiments}
 	\vspace{+2mm}
\end{table*}

With ordinary sharing pattern, we can measure the difference of any two sets of channels $\c_1$ and $\c_2$ with $\ell_{1}$ distance $d(\c_1,\c_2) = \norm{\c_1-\c_2}_1$. Ideally, if the distance $d(\c_1,\c_2)$ is smaller, the representation of channels  $\c_1$ and $\c_2$ should be closer and vice versa. For this reason, we pre-defined a threshold distance $m$, if $d(\c_1,\c_2) < m$, $\c_1$ and $\c_2$ should have similar representation, otherwise, they should be different. With this formulation, we can obtain a conditional distribution by $\blambda$:
\begin{equation}\label{conditional_distribution}
P(\c_i|\c_k) = \frac{\mbox{exp}(\blambda_i^T\blambda_k/\tau)}{\mbox{exp}(\blambda_i^T\blambda_k/\tau) + \sum_{j=1}^{n} \mathds{1}_{[j \neq k]}  \mbox{exp}(\blambda_i^T\blambda_j/\tau)}
\end{equation}
which represents the likelihood that $d(\c_i,\c_k) < m$, and $\tau$ is a temperature parameter that controls this loss's concentration level.
With Eq.\eqref{eqcz}, we proposed a non-parametric regularization term for the simplex-net to distinguish different channel width under the same architecture(operation).  The learning objective is then to optimize the conditional probability of similarity for different channels, which can be expressed by the negative log-likelihood as follows:
\begin{equation}\label{non_parametric_term}
r_\c(\bsigma)  =  - \mbox{log}~P(\c_i|\c_k)
\end{equation}
Hence, the objective function for simplex-net is as follows: 
 \begin{equation}\label{final_loss}
 L(\bsigma) = \mathcal{L}_{train}(\bm{a},\bm{c},\widetilde{\btheta}(\bsigma;\bTheta,\a,\c)) + \alpha \cdot r_\c(\bsigma),
\end{equation}
where $\alpha$ is the trade-off hyperparameter (we find $\alpha=1$ suffices in our experiments). The overall training procedure for our $K$-shot supernets and simplex-net is summarized as Algorithm \ref{supp_algorithm2}.

\section{Experimental Results}
\subsection{Search on ImageNet}
\textbf{Dataset.} We perform the architecture search on the large-scale dataset ImageNet (ILSVRC-12)~\cite{russakovsky2015imagenet}, which contains 1.28M training images from 1000 categories. Specifically, following \cite{guo2020single}, we randomly sample 50K images from the training set as the local validation set, with the rest images used for training. Finally, we report the accuracy of our searched architecture on the test dataset (which is the public validation set of the original ILSVRC2012 ImageNet dataset). All experiments are implemented with PyTorch \cite{pytorch} and trained on 8 NVIDIA Tesla V100 GPUs.

\textbf{Search space.} In this paper, we implement the joint search of operations and channels for the fine-grained NAS. We follow the same search space as SPOS \cite{guo2020single}, which has 21 to-be-searched layers and 3 fixed layers. For channels, we search for all 24 layers.  For operation search, we adopt the same macro search space as other one-shot methods for a fair comparison. Concretely, to construct the supernet, we leverage the MobileNetV2 inverted bottleneck \cite{mobilenetv2} with an optional squeeze-and-excitation (SE) \cite{hu2018squeeze} module. For each search block, the convolutional kernel size is determined within {3,5,7} with the expansion ratio selected in {3,6}, and each block can choose to whether use SE module or not. An additional identity block is also attached for a flexible depth search. Moreover, to accommodate the channel search, the width coefficient of each operation is selected within $\{0.2,0.4,0.6,0.8,1.0\}$ for all 24 layers. As a result, the overall search space amounts to be $5^{24}\times13^{21}$.

\textbf{Supernet training.} We adopt $b_s$ and $\tau$ as 16 and 0.3 for Eq.\eqref{iterative_supners} and Eq.\eqref{conditional_distribution}, respectively. For training $K$-shot supernets, we follow the same training recipe as \cite{you2020greedynas, guo2020single} for a fair comparison. With a batch size of 1024, the supernets are trained using a SGD optimizer with 0.9 momentum and Nesterov acceleration. The learning rate is initialized as 0.12 and decay with cosine annealing for 120 epochs. For each sampled architecture, we use the same code $\blambda$ (\ie, $1/K$) for the $K$-shot supernets within the first 5 epochs to warm up all the weights. As for the simplex-net, we adopt a two-layer MLP for both encodings (\ie, Architecture and Channel). We empirically find that more layers of MLP do not bring further significant improvement, which infers that a two-layer MLP may be enough for investigating the encodings. Then, we include our proposed simplex-net to learn the customized code $\blambda$ via an alternate iterative procedure. More details of training and search are illustrated in supplementary materials.


\textbf{Retraining.}  For retraining the searched architectures, we follow the same strategy as previous works \cite{mnasnet, you2020greedynas}. Specifically, we adopt a RMSProp optimizer with 0.9 momentum and 1024 batch size. The learning rate is increased from 0 to 0.128 linearly for 5 epochs and then decays 0.03 every 2.4 epochs. Besides, the exponential moving average is also adopted with a decay rate of 0.9999.

\textbf{Performance of searched structures.} We perform the search with 4 different groups of FLOPs constraints; each group is composed of a pair of experiments with/without channel width search. As shown in Table \ref{nas_experiments}, with the same search space, our searched 343M $K$-shot-B achieves 77.2\% on Top-1 accuracy, which surpasses other methods by more than 0.4\%. Moreover, we conduct the joint search of operations and channels; the accuracy is further boosted by 0.2\%$\sim$0.3\%, which validates our method's effectiveness.

\subsection{Search on NAS-Bench-201}\label{sec5.2}
The key challenge of the NAS algorithm lies in the evaluation and ranking reliability of the supernet, which can be reflected by the ranking correlation between the evaluation performances of all architectures on supernets and their actual performances. Hence, for a better evaluation of our methods, 
we tend to use the NAS-Bench-201 \cite{nasbench201} to analyze our supernets and architecture search. NAS-Bench-201 is a NAS benchmark that contains 15625 architectures and provides the train-from-scratch performances of these architectures evaluated on ImageNet-16-120, CIFAR-100, and CIFAR-10. 

\textbf{Searching Results of NAS-Bench-201 and comparison of ranking ability.} In $K$-shot supernets, each architecture can acquire its customized code $\blambda$ to migrate the gap between stand-alone training and joint training on supernets. As a result, our method may lead to a more accurate ranking on different architectures and better performance. As Table \ref{NAS-Bench-201} shows, 
our searched results achieve much higher performance on NAS-Bench-201 set, \eg~0.7\% higher than AngleNet with ImageNet dataset. Moreover, we also calculate the Kendall's Tau correlation coefficients between the validation accuracy on supernet and their ground-truth performances. As in Table \ref{tab:kendall_methods}, we run each method 10 times and report its average Kendall's Tau value. The results show that our method achieves a much higher Kendall's Tau with the proposed $K$-shot supernets, \eg, 7.64\% higher than the Single-path method with the CIFAR-10 dataset, which indicates that our $K$-shot supernets estimates the validation accuracy better, and thus searches for the good architectures more accurately.


\begin{table}[t]
	\centering
	\small
	\scriptsize
\caption{Searching results (mean$\pm$std) on NAS-Bench-201 dataset.}
\setlength{\tabcolsep}{0.9mm}{
\begin{tabular}{c||ccc}
		\hline
		Method & CIFAR-10 & CIFAR-100 & ImageNet-16 \\	
		\hline	
		GDAS \cite{dong2019search} & 93.52 $\pm$ 0.15 & 67.52 $\pm$ 0.15 & 40.91 $\pm$ 0.12 \\
		DARTS- \cite{chu2021darts} & 93.80 $\pm$ 0.40 & 71.36 $\pm$ 1.51 & 45.12 $\pm$ 0.82 \\
		SPOS \cite{guo2020single} & 93.67 $\pm$ 0.26 & 69.83 $\pm$ 0.21 & 44.71 $\pm$ 0.17 \\ 
		AngleNet \cite{hu2020angle} & 94.01 $\pm$ 0.37 & 72.96 $\pm$ 0.26 & 45.83 $\pm$ 0.19 \\
		$K$-shot NAS & 94.19 $\pm$ 0.16 & 73.45 $\pm$ 0.05 & 46.53 $\pm$ 0.11 \\ \hline
		optimal & 94.37 & 73.51 & 47.31 \\ 
		\hline
\end{tabular}}
\label{NAS-Bench-201}
\end{table}

\begin{table}
	\centering
	\caption{The comparison of Kendall's Tau w.r.t different methods on NAS-Bench-201. }  
	\label{tab:kendall_methods}
	\small
	\setlength{\tabcolsep}{0.35mm}{
	\begin{tabular}{c||ccc}
		\hline
		Method & CIFAR-10 & CIFAR-100 & ImageNet-16 \\	
		\hline	
		SPOS \cite{guo2020single} & 55.00\% & 56.00\% & 54.00\% \\ 
		AngleNet \cite{hu2020angle} & 57.48\% & 60.40\% & 54.45\% \\
		$K$-shot NAS & 62.64\% & 61.22\% & 56.33\%\\
		\hline
	\end{tabular}}
\end{table}

\begin{figure}[t]
	\centering
	\includegraphics[width=0.7\linewidth]{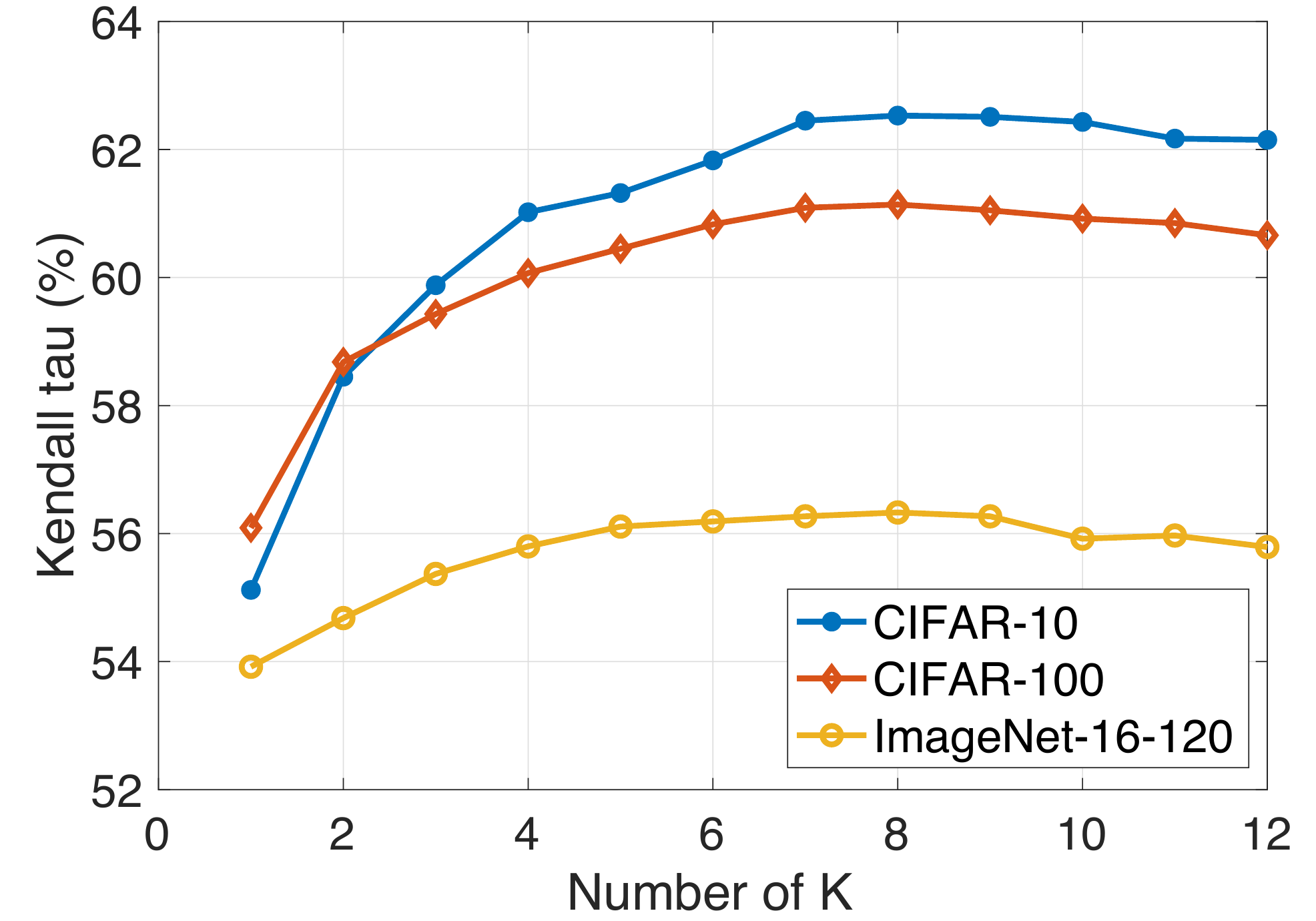}
	\caption{Top-1 accuracy of searched models on ImageNet dataset by different methods with the increasing of search numbers. }
	\label{kendall_tau_K}
 	\vspace{+2mm}
\end{figure}

\textbf{Comparison between number of $K$ for supernets.} 
Since the proposed supernets impose a rank-K approximation for architectures within search space, the number of $K$ is a critical hyperparameter to discuss. Figure \ref{kendall_tau_K} shows the searching result for different number $K$. As we can see clearly, the Kendall's Tau coefficient of $K$-shot supernets benefits from the increase of $K$ (especially when $K$ is small), demonstrating our proposed $K$-shot supernets can effectively bridge the gap between supernets and true performance. Such observation further verified our hypothesis that the number of $K$ determines the approximation level between supernets and true performance, and thus larger $K$ induces a more accurate ranking for sub-networks. As a result, the ranking ability of supernets will be closer to the Oracle optimal with a larger number of $K$, \eg~$K=8$. Moreover, the performance tends to be optimal when $K$ selected from $[6-10]$ but decreases afterward; this is because the larger $K$ induces more weights for $K$-shot supernets, and thus $K$-shot supernets need longer training time.  See more experiments about training $K$-shot supernets in supplementary materials.

\begin{figure}[t]
	\centering
	\includegraphics[width=0.9\linewidth]{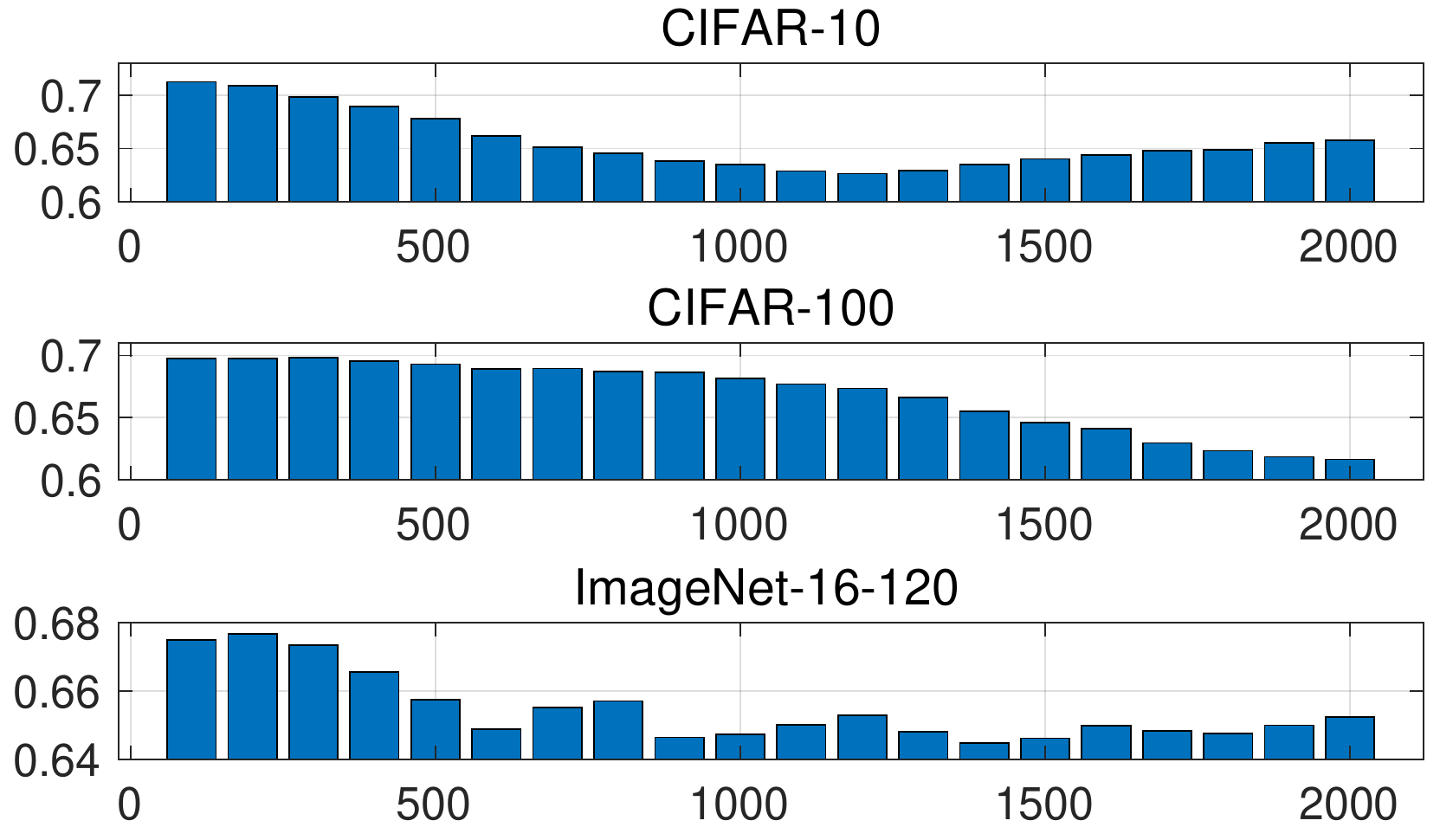}
	\caption{The  Kendall’s Tau coefficient of $K$-shot supernets for high-performance architectures. The abscissa value indicates the number of architectures in high-performance group $\a_{high}$.}
	\label{kendall_tau_high}
 	\vspace{+2mm}
\end{figure}

\textbf{The ranking ability for the high-performance architectures.} We have used the  Kendall's Tau to show the powerful ranking ability of our $K$-shot supernet. However, a high Kendall's Tau score cannot guarantee that the best architecture has the highest ranking. In NAS, we care more about the optimal architectures with the given computational budgets. Intuitively, a natural question comes to \emph{whether the proposed $K$-shot supernets enable to rank high-performance architectures more accurately? } With this aim, we divide the architectures into the high-performance group and low-performance group based on their stand-alone accuracy. Now, we can reformulate Kendall's Tau score as follows: ``For any pair ($\a_{high}$, $\a_{low}$), it is said to be concordant if $\a_{high}$ better than $\a_{low}$, otherwise, it will be called disconcordant". The new Kendall's Tau for evaluating the high-performance architecture can be calculated by $(concordant - disconcordant) / all pairs $. As shown in Figure \ref{kendall_tau_high}, we show Kendall's Tau with different dividing ranks. In detail, with the proposed $K$-shot supernets, the top performance architectures can be more accurately ranked, \eg, the top 100 architectures can be ranked with Kendall's Tau coefficient of 0.68 while 0.56 for the overall architectures. More details about Kendall's Tau and its experiments are formulated in the supplementary materials.


\begingroup
\setlength{\tabcolsep}{2pt}
\begin{table}[t]
	\caption{The performance gain of each part in $K$-shot supernets with 420M FLOPs on MobileNet search space.}
	\label{MCT_NAS_analysis}
	\centering
    \small{
	\begin{tabular}{c|c|c|c|c|c||cc} \hline
		\#&$K$-shot & One-shot &  simplex-net & Channel & $r_c$ & Top-1 & Top-5 \\ \hline
		1& & \checkmark& & & &77.0\% & 93.4\%\\ 
		2& & \checkmark&& \checkmark& &77.1\% & 93.4\%\\ 
	    3&\checkmark& & & & & 77.2\% & 93.5\%\\ 
		4&\checkmark& & \checkmark& & & 77.5\% & 93.6\%\\ 
		5&\checkmark& & \checkmark& \checkmark& & 77.6\% & 93.6\%\\ 
		6&\checkmark& & \checkmark& \checkmark& \checkmark& 77.9\% & 93.8\%\\ \hline
	\end{tabular} }
 	\vspace{+2mm}
\end{table}
\endgroup

\subsection{Ablation Studies}\label{exp:ablation_studies}

\textbf{Effect of each proposed techniques in $K$-shot supernets.} 
As shown in Table \ref{MCT_NAS_analysis}, we conduct the ablations about the proposed $K$-shot supernets. We evaluate the performance with different combinations of these strategies on ImageNet and report their Top-1 accuracy in Table \ref{MCT_NAS_analysis}. With 420M FLOPs budget, we re-implement Single-path \cite{guo2020single}, and it  achieves 77.02\% top-1 accuracy on ImageNet dataset in Table \ref{MCT_NAS_analysis}(\#1). Besides, if we joint search operations and widths with a one-shot strategy, it achieves 77.14\% top-1 accuracy(\#2). When searching with $K$-shot supernets but with the coefficient vector fixed as $1/K$ (\#3), the search result only improves by 0.12\%. However, by comparing \#3 and \#4, if we use simplex-net to generate the customized code for $K$-shot supernets, we observe that the simplex-net and $K$-shot supernets are indeed helpful for the search result. Moreover, by comparing \#5 and \#6, if we promote our method to the channel width search and incorporate the non-parametric regularization, we can further boot our search result with 0.28\%.

\begin{figure}[t]
	\centering
	\includegraphics[width=1.0\linewidth]{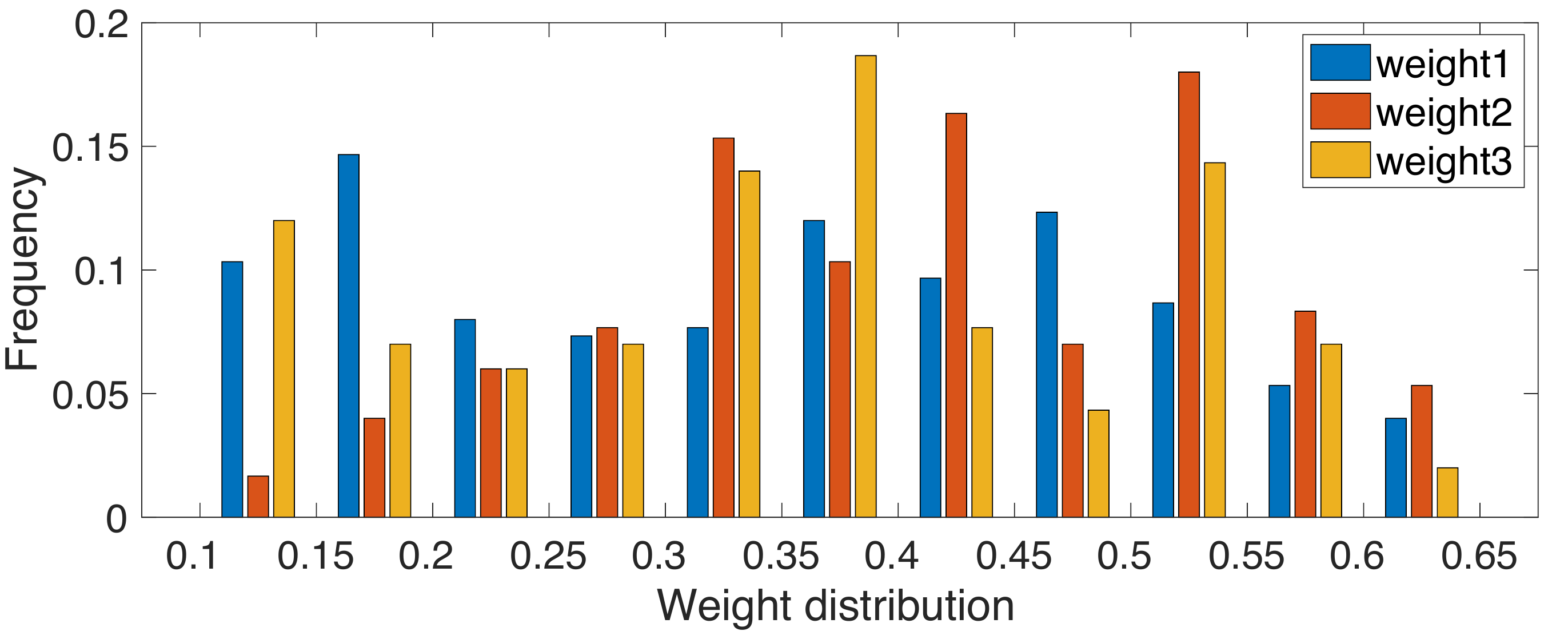}
	\vspace{-6mm}
	\caption{Histogram of weight distribution with $K=3$ of 1000 sub-networks during evolutionary search \wrt 420M FLOPs on ImageNet dataset.}
	\label{weight_tensor}
\end{figure}

\textbf{Visualization of customized code $\blambda$.} In our method, we aim to encourage different architectures to have specialized weights, which are assumed to be represented with a convex combination of $K$-shot supernets via the customized code $\blambda$. In this way, the code actually reflects the preference over all supernets for each path. To examine this \textit{preference}, we record the customized codes $\blambda$ of all searched 1K architectures (420M FLOPs) during the evolutionary searching on ImageNet dataset with $K=3$, then we show their histogram in Figure \ref{weight_tensor}. Concretely, the value of these 3 customized codes $\blambda$ distributes evenly within a large range from $0.1$ to $0.65$, which indicates our method can allocate customized code in a wide range for different architectures. Besides, the frequency of these 3 customized codes $\blambda$ varies accordingly for different architectures, which implies that these 3 customized codes are learned to reflect the different potential of architectures by various preferences.  More visualization of customized code with detailed explanations is illustrated in supplementary materials.

\begin{table}[t]
	\centering
	\scriptsize
	\caption{Comparison of $K$-shot supernets with baseline methods on ImageNet dataset \wrt 330M and 420M FLOPs budget.}
	\begin{tabular}{c|c|c||c|c}
	\hline
	Method & FLOPs & Accuracy(\%) & FLOPs & Accuracy(\%) \\ \hline
	One-shot-NAS & \multirow{4}*{330M} & 76.2 & \multirow{4}*{420M}  & 77.0 \\ 
	Random $K$-shot-NAS & & 76.2 & & 77.1\\ 
	fixed K-shot-NAS & & 76.4 & & 77.2 \\ 
	$K$-shot-NAS & & 77.2 & & 77.6 \\ \hline
	\end{tabular}	
	\label{ablation_simplexnet_NAS}
\end{table}

\textbf{Effect of simplex-net.} We introduce two baseline methods to investigate the effect of simplex-net, namely, \textit{fixed K-shot supernets}: the customized code $\blambda$ is fixed with $1/K$ for all architectures. \textit{Random K-shot supernets}: we randomly initial parameters $\blambda$ and never change it during training for 10 times, and record the average searched performance. From Table \ref{ablation_simplexnet_NAS}, our method achieves much higher performance than baseline methods; while randomly or fixed (\eg, $1/K$) assign $\blambda$ for the introduced $K$ times of supernets nearly do not affect the accuracy performance, which indicates that our simplex-net can effectively boost the performance by providing customized code $\blambda$ for each architecture.

\begin{figure}[t]
	\centering
	\includegraphics[width=0.8\linewidth]{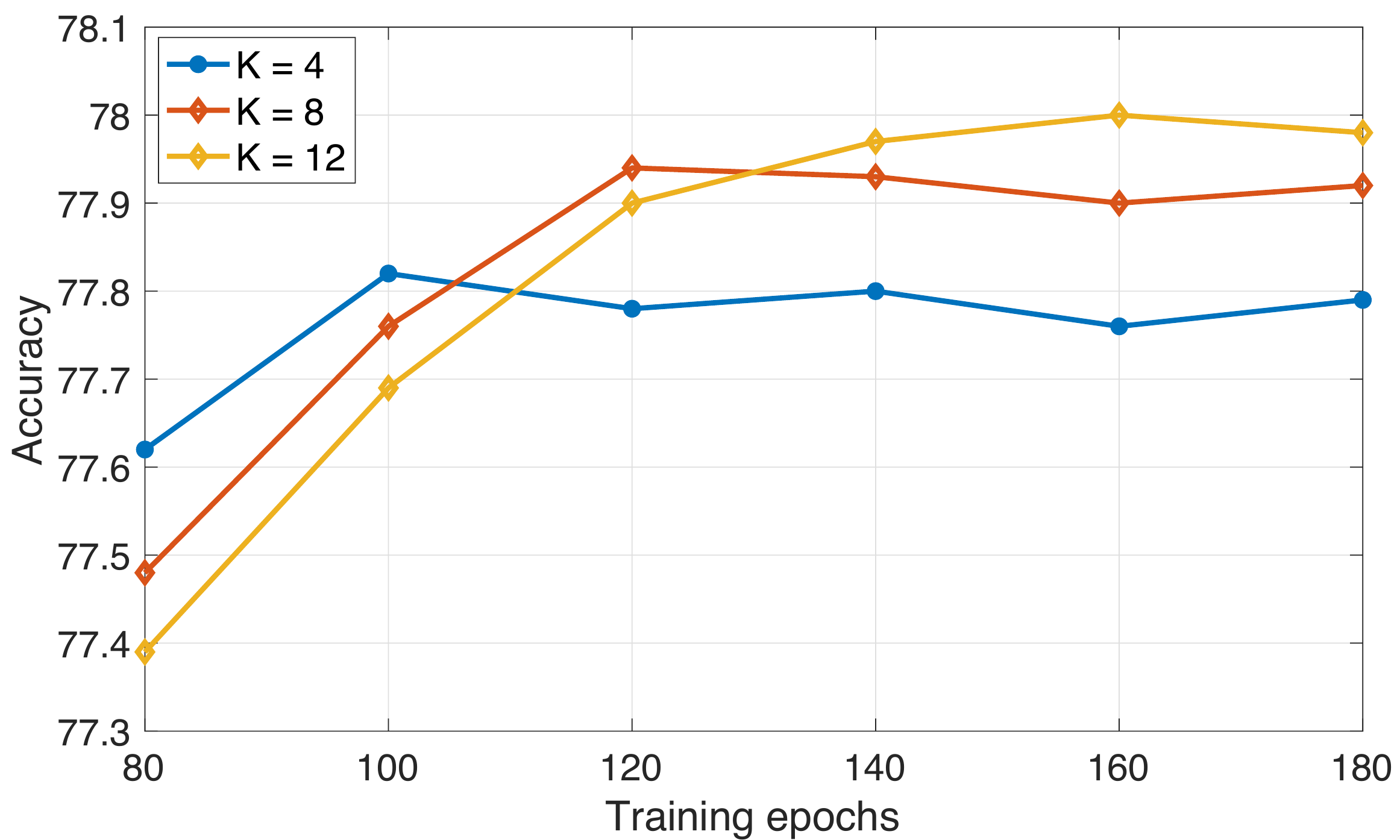}
	\caption{Top-1 accuracy of searched models on ImageNet dataset \wrt different $K$ and supernet training epochs. }
	\label{Trade_off}
 	\vspace{+2mm}
\end{figure}

\textbf{Tradeoff between Training Epochs and Number of $K$.}
In K-shot NAS, we leverage $K$ supernets and only introduce negligible training cost and GPU memory as shown in Table \ref{time_cost}. However, with ($K$-1) times more parameters than the ordinary one-shot NAS, the $K$-shot supernets may become hard to be optimized. To investigate the trade-off between training epochs and the number of $K$, we implement the search with different $K$ and supernet training epochs on ImageNet dataset. From Figure \ref{Trade_off} we can see that, on the one hand, with a larger $K$, supernets need to be trained with more epochs for a more accurate ranking of subnets. On the other hand, with a tight training budget, smaller $K$ may induce better training results. As a result, for the trade-off between training cost and the number of $K$, we simply use $K=8$ in our method.












\section{Conclusion}
In this work, we discussed the drawbacks of the one-shot NAS and proposed a novel $K$-shot weight sharing strategy for $K$-shot supernets. Besides, we introduced a simplex-net which learns context information from architectures/channel encoding vectors, and generate the customized code that cooperate with the $K$-shot supernets to approximate the optimal stand-alone weights. Furthermore, we designed a regularization term that helps the simplex-net generate different code for different channel width. Extensive experiments have been implemented on the large-scaled ImageNet dataset and NAS-Bench-201 to show the superiority of our propose method to other state-of-the-art NAS methods.


\subsubsection*{Acknowledgments}
This work is funded by the National Key Research and Development Program of China (No. 2018AAA0100701) and the NSFC 61876095. Chang Xu was supported in part by the Australian Research Council under Projects DE180101438 and DP210101859. Shan You is supported by Beijing Postdoctoral Research Foundation.

\bibliography{reference}
\bibliographystyle{icml2021}

\onecolumn

\appendix
The supplementary materials are organized as follows. In Appendix \ref{A1}, we illustrate the details of the training recipe. In Appendix \ref{A2}, we provide the implementation details of the evolutionary search. In Appendix \ref{A3}, we illustrate the definition of rank correlation coefficients and report the corresponding experimental results. We compare our $K$-shot NAS with an ensemble of $K$ supernets in Appendix \ref{A4}. We report more visualizations of customized code $\blambda$ in Appendix \ref{A5}. In Appendix \ref{A6}, we provide a detailed algorithm flow of iterative training with $K$-shot NAS. In section \ref{A7}, we report the ablations of iterative training of $K$-shot NAS.  Finally, we show the visualization of searched architectures in Table \ref{tbl:sota}.


\section{Details of Training Recipe} \label{A1}
In this section, we present the supernet training and train from scratch details with the proposed $K$-shot NAS \wrt~ImageNet and NAS-Bench-201 datasets. In general, we adopt $m=16$, $a=1$, $\tau=0.3$ for all experiments. Besides,  we leverage an iterative training strategy for supernets and simplex-net, and details are illustrated in  Algorithm \ref{supp_algorithm2}.

\textbf{Supernet training with $K$-shot NAS.} For ImageNet dataset, we follow the same strategy as \cite{you2020greedynas, guo2020single}. In detail, we adopt a batch size of 1024; supernets are trained via a SGD optimizer with 0.9 momentum and Nesterove acceleration. The initial learning rate is set to 0.12 with a cosine annealing strategy, which decays 120 epochs. For each sampled architecture, we use the same code $\blambda = 1/K$ for the $K$-shot supernets within the first 5 epochs to warm up all supernets. Then we include the proposed simplex-net to learn the customized code $\blambda$ through an alternate iterative procedure, which is presented in Algorithm \ref{supp_algorithm2}. Especially, when training simplex-net while fix supernets, we divide the image batch into $m$ groups, with each group shares the same architectures. Therefore, the input batch size for simplex-net will be as $m>1$, which promotes a better optimization for simplex-net with Eq.\eqref{iterative_supners}. Besides, for NAS-Bench-201 dataset, we simply follow the same training strategy provided in \cite{nasbench201} for supernet training.

\textbf{Retraining of searched architectures.} To train the searched architectures from scratch, we use the same retraining strategy as \cite{guo2020single} for a fair comparison. In detail, we train the searched architecture from scratch with RMSProp optimizer and 0.9 momentum; the learning rate is increased from 0 to 0.128 linearly for 5 epochs and then decays 0.03 every 2.4 epochs. Besides, the exponential moving average is adopted with a decay rate set to 0.9999.

\textbf{Training of $K$-shot NAS.} With Eq.\eqref{approxmation_weights}, we merge the weights of all supernets $\theta_k$ to  $\widetilde{\btheta}=\sum_{k=1}^K\lambda_{k}\btheta_k$, then the merged weights $\widetilde{\btheta}$ are optimized as same as One-shot NAS, as formulated in Eq. \eqref{eqx}. Therefore, we keep almost the same budgets as One-shot NAS while training all $K$ supernets simultaneously. The extra gradient calculation cost of $\theta_k$ is small, given a scalar value $\lambda_k$.

\section{Details of Evolutionary Search} \label{A2}
Since the search space is enormous (\eg, $5^{24}\cdot13^{21}$ for joint searching operations and channel width with MobileNetV2 search space), to boost the search efficiency, we leverage the multi-objective NSGA-II \cite{deb2002fast} algorithm for evolutionary search, which is easy to integrate hard FLOPs constraint. We set the population size as 50 and the number of generations as 20, which amounts to 1000 paths in our method. To perform the search, we randomly select a group of architectures within the target FLOPs. In each iteration, the top 20 architectures with the highest accuracy are selected as parents to generate new architectures via mutation and crossover. We evaluate each architecture by the inference with the weights from $K$-shot supernets with Eq.\eqref{search_channels} and record its accuracy. In each generation, the top 20 architectures with the highest accuracy are selected as parents to generate new architectures via mutation and crossover. After the search, we only retrain the architecture with the highest accuracy from scratch and report its performance.

Note that batch normalization (BN) layers are incorporated in most operations (\eg, 3 by 3 inverted residual in MobileNetV2 search space). However, due to the varying channel width, the mean and variance in BN layers are unsuitable for all width. Therefore, we simply use the mean and variance in batches instead, and we set the batch size to 2048 to induce an accurate estimation of mean and variance.

\section{Details of Rank Correlation Coefficient} \label{A3}
\subsection{Defination of kendall's Tau} \label{kendall_tau}
In section \ref{sec5.2}, we implement the search on NAS-Bench-201 and examine the result with Kendall's Tau coefficient.  We rank the evaluation results on various validation sets with 15625 architectures and 50,000 samples. Indeed, the Kendall Tau correlation is used to evaluate the $K$-shot supernet's ranking ability with the pairwise ranking performance. With any pair ($\r_i,\r_j$) and ($\s_i,\s_j$), if we have either both $\r_i>\r_j$ and
$\s_i>\s_j$, or both $\r_i<\r_j$ and $\s_i<\s_j$, these two pairs are considered as \textbf{concordant}. Otherwise, it is said to be \textbf{disconcordant}. Therefore, the Kendall Tau can be formally defined as
\begin{equation}
    K_{\tau} = \frac{n_{\text{con}} - n_{\text{discon}}}{n_{\text{all}}},
    \label{eq:appendix:rank:kendall-tau-var}
\end{equation}
where $n_{\text{con}}$ and $n_{\text{discon}}$ indicates the number of concordant and disconcordant pairs, and $n_{\text{all}} = \mathbb{C}_n^2$ is the total number of pairs.

\textbf{Kendall's Tau for "The ranking ability for the high-performance architectures" in Section \ref{sec5.2}.} To calculate the Kendall's Tau between high-performance architectures and others. We need to divide the architectures into a high-performance group and a low-performance group with the provided true ranks. As a result, for any pair ($a_{high},a_{low}$) during calculation of Eq.\eqref{eq:appendix:rank:kendall-tau-var}, $a_{high}$ and $a_{low}$ are from the high-performance group and the low-performance group, respectively.

\subsection{Other rank correlation coefficients}
To evaluate the search results of $K$-shot NAS, we also introduce two more correlation coefficients, \ie, Spearman rho \cite{pirie2004spearman}, and Pearson \cite{Nahler2009Pearson}. Spearman rho correlation coefficient is the Pearson correlation coefficient between random variable $\r$ and $\s$, \ie
    \begin{equation}
        \rho_S = \frac{\operatorname{cov}(r, s)}{\sigma_r \sigma_s},
        \label{eq:appendix:rank:spearman-rho-var}
    \end{equation}
 where $\operatorname{cov}(\cdot, \cdot)$ is the covariance of two variables, and $\sigma_r$ and $\sigma_s$ are the standard deviations of $r$ and $s$, respectively. Since the ranks are integers in our experiments, the Eq.\eqref{eq:appendix:rank:spearman-rho-var} can be fulfilled more efficiently with:
     \begin{equation}
        \rho_S = 1 - \frac{6 \sum_{i=1}^{n} (\boldsymbol{r}_i - \boldsymbol{s}_i)^2}{n (n^2 - 1)},
        \label{eq:appendix:rank:spearman-rho-vector}
    \end{equation}

Where $n =15625$ indicates the number of overlapped elements between $r$ and $s$.

\textbf{More experiments of ranking correlations of $K$-shot supernets on NAS-Bench-201.} As shown in Table \ref{rank_correlations}, we report the experiments with more ranking correlations of $K$-shot supernets for a detailed analysis of our method.
\begin{table}[H]
	\centering
	\caption{The ablations of $K$-shot NAS \wrt~more ranking correlations.}
	\label{rank_correlations}
	\begin{tabular}{c|c|c|c}
	\hline
	Method& CIFAR-10 & CIFAR-100 & ImageNet-16-120 \\ \hline
	Kendall's Tau & 0.63 & 0.61 & 0.56 \\
	Spearman rho & 0.81 & 0.80 & 0.73\\ 
    Pearson & 0.92 & 0.84 & 0.76 \\ \hline
	\end{tabular}
\end{table}

\section{The Comparison between $K$-shot Supernets and Ensemble of $K$ Supernets} \label{A4}
Since our $K$-shot NAS proposes to learn the customized code for each subnet with $K$ supernets. In this way, one natural question comes to \textit{How does $K$-shot NAS compare to directly ensemble $K$ supernets, \ie, independently train $K$ times of supernets and ensemble their results?} With this aim, with $K$ supernets and ImageNet dataset, we compare the proposed $K$-shot NAS with two baseline methods, \ie, $\mbox{Ensemble}_{max}$: we output the result from the supernet with maximum Top-1 probability. $\mbox{Ensemble}_{avg}:$ we average the output result from $K$ different supernets. We compare our methods with these two baselines in terms of search efficiency and accuracy performance, as shown in Table \ref{ensemble_K}.\footnote{Since the ensemble of $K$ cost much more computation resolutions than our method, we only implement the search with $K = 2$ for ensemble method.} In detail, accuracy performance benefit from ensemble of $K$ supernets, but it also puts a huge burden on the training cost and GPU usage. Nevertheless, our proposed $K$-shot NAS can efficiently boost the search results while introducing a negligible additional computation budget.
\begin{table}[H]
	\centering
	\caption{The comparison of $K$-shot supernets and ensemble of $K$ supernets. The $3$-th 
	column indicates the training cost (GPU hours) of 1 epoch. The $4$-th 
	column reports the GPU usage with the batch size 64 for each GPU.}
	\label{ensemble_K}
	\begin{tabular}{c|c|c|c|c}
	\hline
	Number of $K$ & method &  Training cost (h) & GPU usage (G) & Top-1 (\%) \\ \hline
	K = 1 & one-shot NAS & 2.27 & 8.7  & 77.12 \\
	K = 2 & $\mbox{Ensemble}_{max}$ & 4.54 & 17.5  & 77.58\\ 
	K = 2 & $\mbox{Ensemble}_{avg}$ & 4.55 & 17.6 & 77.64 \\ 
	K = 2 & $K$-shot NAS & 2.28 & 8.8 & 77.56\\
	K = 4 & $K$-shot NAS & 2.32 & 8.9 & 77.79 \\
	K = 8 & $K$-shot NAS & 2.37 & 9.2 & 77.92 \\ \hline
	\end{tabular}
\end{table} 


\section{More Experiments of Visualization of Customized Code $\blambda$} \label{A5}
\begin{figure}[H]
	\centering
	\includegraphics[width=0.98\linewidth]{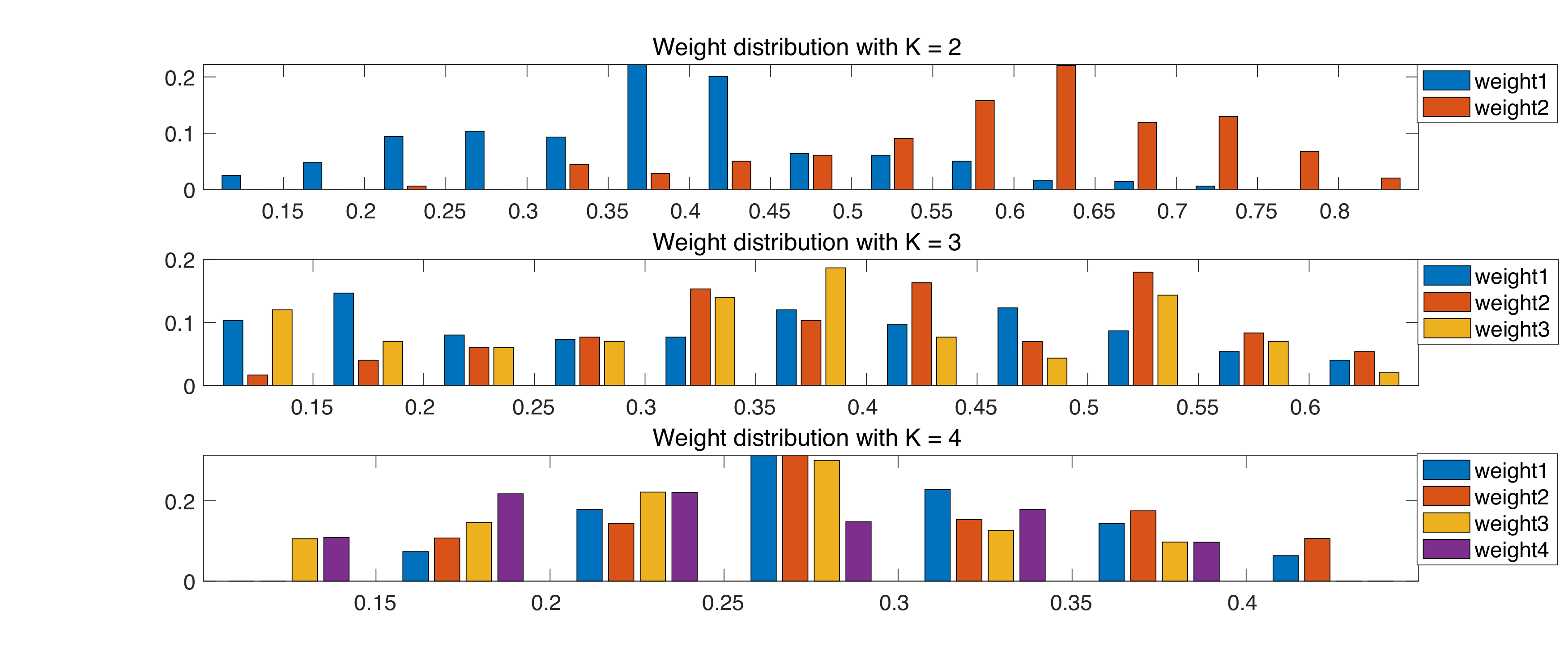}
	\caption{Histogram of weight distribution with different $K$ of 1000 sub-networks during evolutionary search \wrt~420M FLOPs on ImageNet dataset.}
	\label{weight_tensor_addition}
\end{figure}

For intuitively understanding, we visualize the proposed customized code $\blambda$ with 1000 paths searched with Imagenet dataset with 420M FLOPs budget \wrt~different number of $K$. From Figure \ref{weight_tensor_addition}, we show the customized code with $K = 2,3,4$. Concretely, the customized code distributes evenly within a large range \wrt~different $K$, which indicates our method can customize the code for different architectures and promotes to distinguish their performance accurately.

\section{Details of Algorithms Flow with the Iterative Training of $K$-shot NAS} \label{A6}
With $K$-shot NAS, we propose an iterative training method for $K$-shot supernets and simplex-net. For intuitively understanding, we provide a detailed algorithm flow about this process as shown in Algorithm \ref{supp_algorithm2}.
\begin{algorithm}[H]
    \caption{Iteratively training of $K$-shot supernets and simplex-net}
    \label{supp_algorithm2}
    \SetAlgoLined
    \KwIn{$K$-shot supernets with weights $\bTheta$, simplex-net $\pi$ with weights $\bsigma$, maximum training epochs $T$.}
    Init current batch $\tau$ = 0; \\
    \While{$\tau$ $\leq T$}{
        \uIf{ $\tau~\%~2 == 0$}{
            randomly sample an subnets with architecture $\a$; \\
            make one-hot architecture parameters of $\a$; \\
            calculate the customized code $\blambda$ with architecture parameters; \\
            train $K$-shot supernets with weights of $\bTheta\blambda$ ;}
        \Else{ 
        randomly sample $m$ groups of architectures $\a$; \\
        make a batch ($m$) of one-hot architecture parameters with $\a$; \\
        calculate the customized code $\blambda$ with architecture parameters; \\
        calculate regularization term $r_c$ with Eq.\eqref{non_parametric_term};\\
        forward and backward for updating weights $\bsigma$ of simplex-net $\pi$;
        }
    }
    \KwOut{The architecture $\a^*$ searched with Eq.\eqref{search_channels}.}
\end{algorithm}


\section{Ablations of Iterative Training of $K$-shot NAS} \label{A7}
Since the batch size of simplex-net is inversely proportional to the batch size of a single architecture for $K$-shot supernets, we adopt an iterative training strategy to balance the training of $K$-shot supernets and simplex-net in our method. To explore the effectiveness of $K$-shot NAS, we also implement the search by jointly optimizing $K$-shot supernets (\eg, $K=8$) and simplex-net \wrt~different group numbers $m$ as formulated in Eq.\eqref{iterative_supners}.
\begin{figure}[H]
	\centering
	\includegraphics[width=0.5\linewidth]{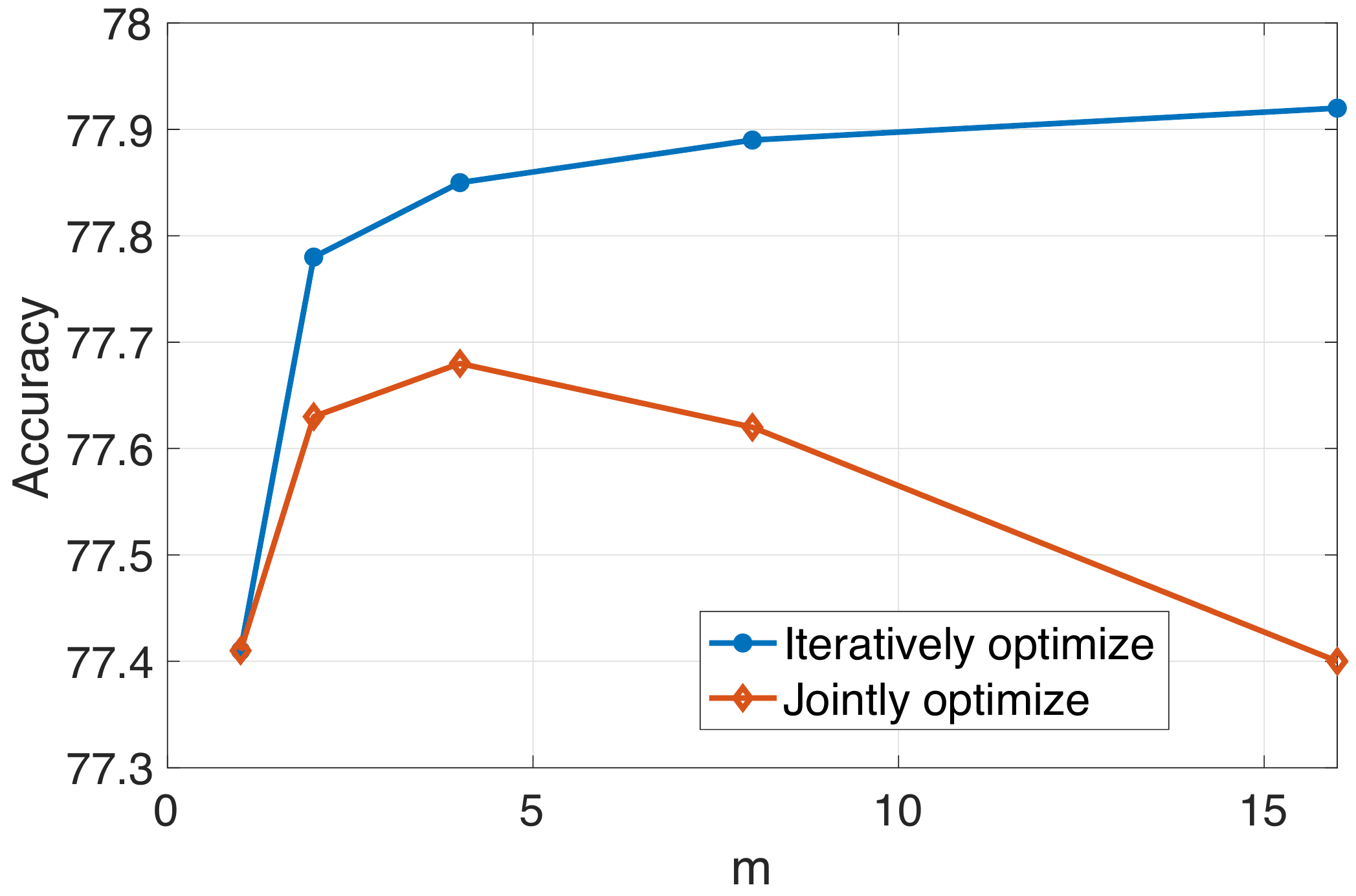}
	\caption{Top-1 accuracy of searched models on ImageNet dataset by different methods with the increasing of group numbers $m$. }
	\label{iterative}
\end{figure}
From Figure \ref{iterative}, our algorithm achieves superior performance with an iteratively training strategy. Concretely, the accuracy performance benefits from the increase of $m$ when $m$ selected from $[1,2,4]$, which is because the increase of batch size $m$ for simplex-net promotes the optimization of customized code $\blambda$. However, after $m$ goes beyond $4$, the searched results of the jointly optimizing method drop gradually, which is because a larger $m$ is not suitable for optimizing $K$-shot supernets. For example, by jointly optimizing of $K$-shot supernets and simplex-net, with the 1024 batch size and $m$ is set to 16, the weights of supernets are optimized with batch size $\frac{1024}{16}=64$ for each architecture, which is far from enough to optimize supernets.



\section{Ablation study of $K$ for Figure \ref{kendall_tau_K}} \label{ab}
The statistical results of K in Figure \ref{kendall_tau_K} are reported in Table \ref{KT}. With K increasing from 1 to 12, the Kendall tau is increased by 2\%$\sim$7\% for 3 datasets, which is a significant improvement since K-shot NAS can achieve
better performance due to the more accurate approximation between supernets and true performance.
\begin{table}[h]
	\centering
\caption{Statistical results of Kendall tau \wrt~$K$ in Figure 4.}
\begin{tabular}{c|ccccccc}
		\hline
		$K$& 1 & 2 & 4 & 8 & 12 \\ \hline
		CIFAR-10 & 55.02 $\pm$ 0.27 & 58.23 $\pm$ 0.21 & 60.96 $\pm$ 0.19 & 62.64 $\pm$ 0.17 & 62.13 $\pm$ 0.13 \\
		CIFAR-100 & 56.07 $\pm$ 0.18 & 58.56 $\pm$ 0.17 & 60.03 $\pm$ 0.11 & 60.40 $\pm$ 0.13 & 60.19 $\pm$ 0.09 \\
		ImageNet-16 & 53.97 $\pm$ 0.20 & 54.36 $\pm$ 0.17 & 55.91 $\pm$ 0.14 & 56.33 $\pm$ 0.12 & 55.89 $\pm$ 0.13 \\ \hline
\end{tabular} \label{KT}
\end{table}

\section{Visualization of Searched Architectures} \label{A9}
\begin{figure}
    \centering
    \subfigure[K-shot-NAS-A$^\star$]{\includegraphics[width=0.20\linewidth]{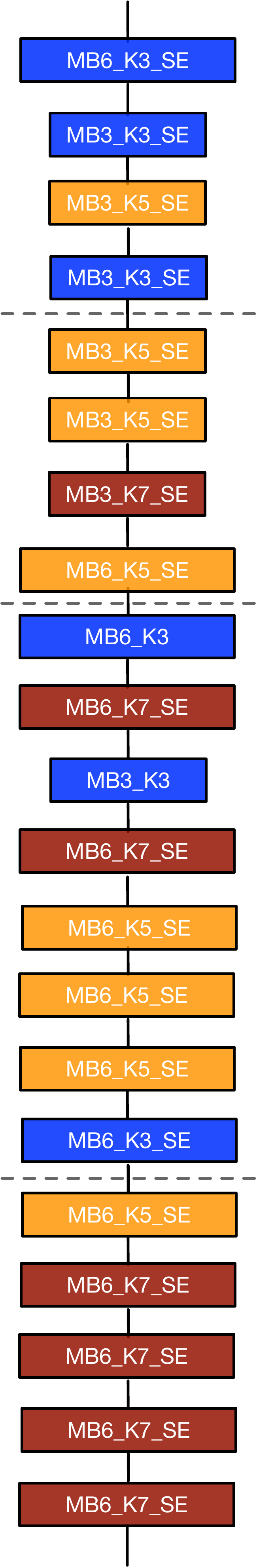}}
    \hspace{5mm}
    \subfigure[K-shot-NAS-B$^\star$]{\includegraphics[width=0.20\linewidth]{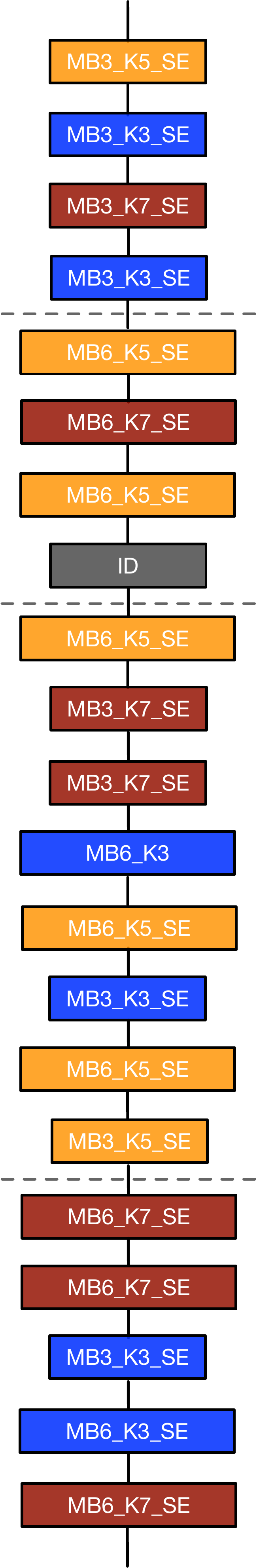}}
    \hspace{5mm}
    \subfigure[K-shot-NAS-C$^\star$]{\includegraphics[width=0.20\linewidth]{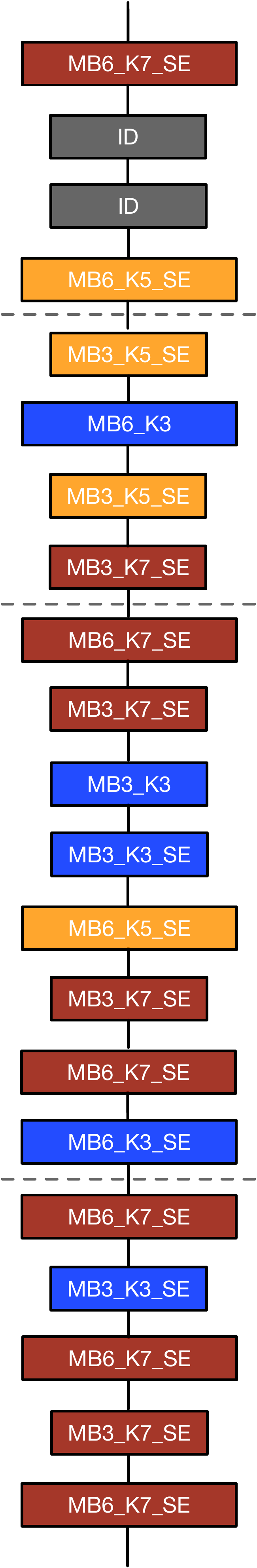}}
    \hspace{5mm}
    \subfigure[K-shot-NAS-D$^\star$]{\includegraphics[width=0.20\linewidth]{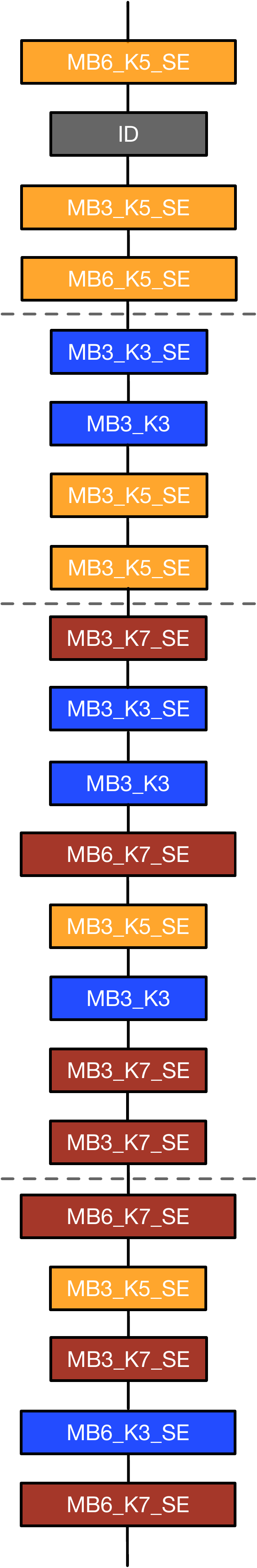}}
    \caption{Visualization of architectures searched by $K$-shot NAS in Table \ref{tbl:sota}.}
    \label{fig:sota_vis_op}
\end{figure}

\begin{figure}[H]
	\centering
	\includegraphics[width=1.00\linewidth]{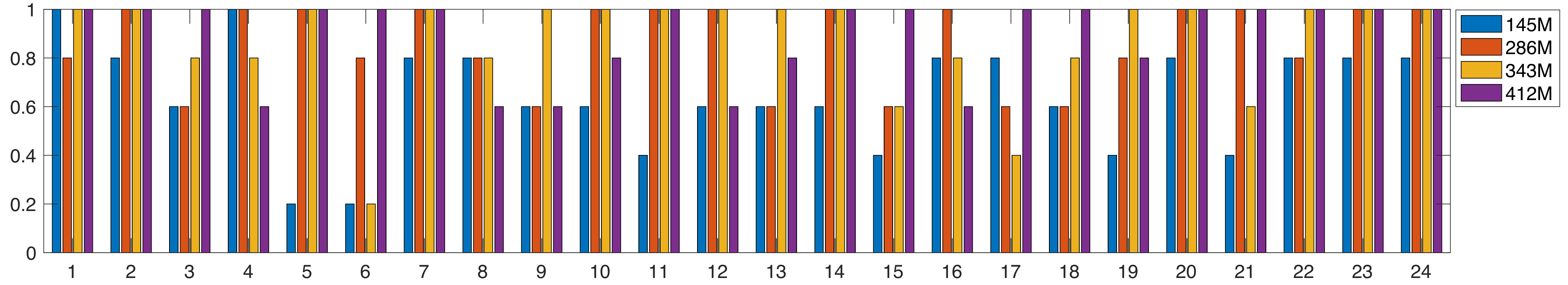}
	\caption{Visualization of channel width searched by $K$-shot NAS in Table \ref{tbl:sota}.}
	\label{fig:sota_vis_channel}
\end{figure}

We visualize the searched architectures (\ie, operations and channel width) with our $K$-shot NAS as Figure \ref{fig:sota_vis_op} and Figure \ref{fig:sota_vis_channel}. In detail, the searched optimal architectures share similar characteristics as follows:
\begin{enumerate}
	\item The optimal architecture generally tends to use more 5$\times$5 or 7$\times$7 convolutions with full channel width at the layers close to the last layer. Significantly, the last layer is always with 7$\times$7 kernel size.
	\item The channel width in the layers close to the input and output generally tends to be fully preserved. Besides, we also obverse that the channel width in the layers with stride 2 also has the full width. 
	\item If the computation budget is insufficient (\ie, 343M and 145M), more ID operations will be used at the layers close to the first layer for searching optimal structures.
\end{enumerate}

Notably, we notice that the 412M searched architecture keeps most of its channels for all layers, which may result from two reasons. First, for operations, we search architectures with different expansion ratios (\ie, 3 or 6), which also determines the width for layers. Second, 412M FLOPs is a relatively large budget, and it thus promotes more channels in each layer to boost the performance.

\end{document}